\documentclass[conference]{IEEEtran}
\IEEEoverridecommandlockouts
\usepackage{cite}
\usepackage{amsmath,amssymb,amsfonts}
\usepackage{algorithmic}
\usepackage{graphicx}
\usepackage{textcomp}
\usepackage{xcolor}
\usepackage{subfig}
\usepackage{amsmath}
\usepackage{amssymb}
\usepackage{booktabs}
\usepackage{multirow}
\usepackage{hyperref}

\usepackage{cuted}

\def\BibTeX{{\rm B\kern-.05em{\sc i\kern-.025em b}\kern-.08em
    T\kern-.1667em\lower.7ex\hbox{E}\kern-.125emX}}
    
\begin{document}
\title{Do as we do: \\Multiple Person Video-To-Video Transfer}

\author{
\IEEEauthorblockN{Mickael Cormier\textsuperscript{3,1}\quad Houraalsadat Mortazavi Moshkenan\textsuperscript{3}\quad Franz Lörch\textsuperscript{1} \quad Jürgen Metzler\textsuperscript{1,2}\quad Jürgen Beyerer\textsuperscript{1,3}}
\\
\IEEEauthorblockA{
    \textsuperscript{1}\textit{Fraunhofer IOSB}, Karlsruhe, Germany; 
    \textsuperscript{2}\textit{Fraunhofer Center for Machine Learning}; \\
    \textsuperscript{3}\textit{Vision and Fusion Lab, Institute for Anthropomatics and Robotics}, \\\textit{Karlsruhe Institute of Technology (KIT)}, Karlsruhe, Germany} 
    \{mickael.cormier, franz.loerch, juergen.metzler, juergen.beyerer\}@iosb.fraunhofer.de, ueefc@student.kit.edu
}

\maketitle

\begin{strip}\centering
\vspace{-0.55in}
\includegraphics[width=\textwidth]{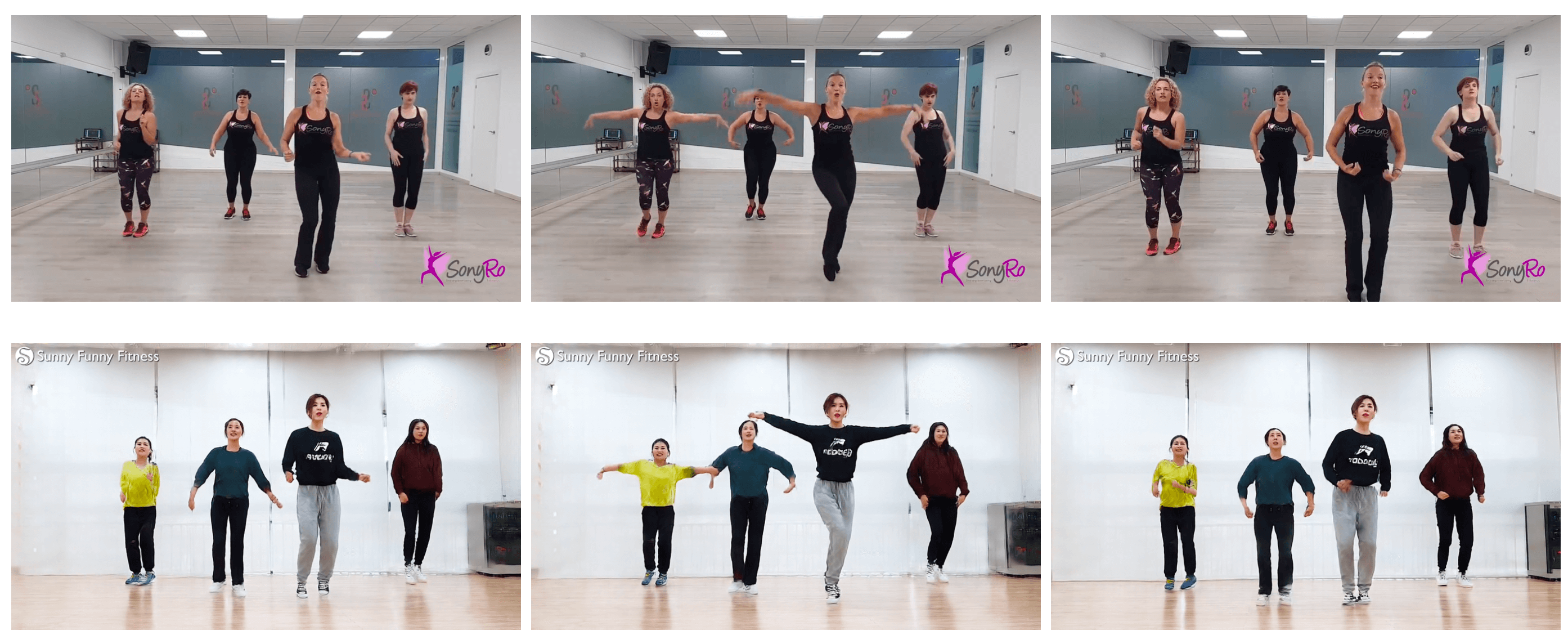}
\vspace{-.3in}
\captionof{figure}{{\bf``Do as we do'' motion transfer:}  given a clip of a team working out~\cite{4person2} (top), and a video of people performing other exercises~\cite{4person1}, our method transfers the workout onto the second team (bottom).}
\label{fig:transfer_1}
\end{strip}
\begin{abstract}
\noindent
Our goal is to transfer the motion of real people from a source video to a target video with realistic results. While recent advances significantly improved image-to-image translations, only few works account for body motions and temporal consistency. However, those focus only on video re-targeting for a single actor/ for single actors. In this work, we propose a marker-less approach for multiple-person video-to-video transfer using pose as an intermediate representation. Given a source video with multiple persons dancing or working out, our method transfers the body motion of all actors to a new set of actors in a different video. Differently from recent "do as I do" methods, we focus specifically on transferring multiple person at the same time and tackle the related identity switch problem. Our method is able to convincingly transfer body motion to the target video, while preserving specific features of the target video, such as feet touching the floor and relative position of the actors. The evaluation is performed with visual quality and appearance metrics using publicly available videos with the permission of their owners. 

\end{abstract}

\section{Introduction}
\vspace{-.1in}

Human Motion analysis is an important topic in the computer vision community. Recent advances in human-related application systems bring new human-centered challenges such as driver behavior recognition~\cite{drive_and_act_2019_iccv}, crowd pose estimation for crowd motion analysis~\cite{golda2019crowdposeestimation} or human action recognition in the dark~\cite{xu2020arid}. However, the CNNs trained for such higher-level tasks require large amounts of annotated data. This data is often challenging to collect and properly annotate. Therefore, synthetic photo-realistic data is often considered as a cost-effective method for augmentation of existing datasets~\cite{3dsemseg_ICCVW17,fabbri2018learning,kviatkovsky2020real}. In this work, we introduce a method for synthesizing real-looking videos of multiple persons dancing side by side and switching places based on real input and target videos.
Based on the Everybody Dance Now work~\cite{chan2019dance} and similar to recent video-to-video translation works~\cite{zhou2019dance,gomes2020,ren2020}, we first extract pose skeletons using a state of the art method~\cite{cao2017realtime,openpose1,openpose2,openpose3}. Since those works only address the video-to-video translation problem for a single person, we extend the simple yet efficient method from~\cite{chan2019dance} to the multiple person transfer problem. We collect online videos from dance workouts with different numbers of persons and perform an ablation study depending on the number of persons. Furthermore, we improve the face generation network by using more accurate face landmarks. Finally, this scenario brings new challenges regarding the pose transfer of each individual in the group. The normalization step is adapted in order to accurately map each subject from the input video to its counterpart in the target. Furthermore, we address the problem of persons switching places by adapting the keypoint correspondence network from~\cite{umer2020self} for tracking each individual in the video.

\section{Related Work}
\label{sec:relatedwork}

Recent breakthroughs in the field of image-to-image translation were recently offered by the introduction of conditional GANs for paired and unpaired images~\cite{Isola_2017,zhu2017unpaired}. Those works were rapidly followed by numerous methods for image and video manipulation. In this section, we review related work for image-to-image translation and appearance transfer.

Wang et al.~\cite{Wang_2018} presented a method to generate high resolution 2048$\times$1024~pixels results from semantic label maps using perceptual loss, a coarse-to-fine generator and a multi-scale discriminator architecture. An approach was proposed in~\cite{lassner2017generative} to generate images of high-resolution using semantic segmentation and texture prediction. Various generative adversarial networks were proposed to increase the visual quality of generated images using labels and texts~\cite{zhu2017your, zhang2017stackgan, yan2016attribute2image, odena2017conditional}. Liu et al.~\cite{liu2020pose} proposed an encoder-decoder for a pose-guided high resolution appearance transfer to a target pose. They use local descriptors with means of progressive local perceptual loss and local discriminators at the highest resolution followed by training of the autoencoder architecture. Zanfir et al.~\cite{zanfir2018human} successfully transferred the appearance of a person in source images to a person in target images while preserving the body outline of the target person, using 3D pose as an intermediate representation. Kundu et al.~\cite{kundu2020} propose a recurrent network for targeting a long-term synthesis of 3D person interactions for long periods of time. Attribute-Decomposed GAN~\cite{men2020controllable} introduces a generative model for controllable person image synthesis, which generates desired human attributes such as pose, head, upper clothes and pants.

Efros et al.~\cite{Efros03} transfers videos based on predicted skeletons introducing the concepts of “Do as I do”, where the images of a target person are generated according to a drivers movement, and “Do as I say” where images of target persons are produced based on imposed commands. 
More recently Zhou et al.~\cite{zhou2019dance} trained a model with a relatively long video of a target person which resulted in the ability to transfer any movements of choice from a reference video to the target person while preserving the appearance of the target person. This model receives a frame of a target person and a pose from the reference as an input and generates the images of the target person in that pose as an output. Wang et al.~\cite{wang2019few} proposed a model for generating images of the targets including humans or scenes that have never been seen previously as a few-shot vid2vid framework. This model generalizes the poses of the reference video to few example images of the target simultaneously. Liu et al.~\cite{liu2019video} proposed a generative adversarial learning-based approach to upper body video synthesis. They use body and facial landmarks of the source person into the target person, followed by the normalization of the upper body landmarks to generate facial features in the target video with spatio-temporal smoothing. Chan et al.~\cite{chan2019dance} proposed a similar approach for motion transfer from a source video to a target video. Their approach consists of two steps of pose encoding and normalization followed by a pose to video translation. Their poses are normalized by evaluating the ankle positions and height of the subjects in order to adapt the size of the source to the target. Their pose to video translation uses a three-step coarse to fine approach, one step explicitly addressing the quality of the generated face, and use of temporal smoothing. Videos for unseen in-the-wild poses are generated in~\cite{ren2020} using data augmentation and unpaired learning to improve generalization of the system for minimizing the domain gaps between testing and training pose sequence. Gomes et al.~\cite{gomes2020} account for pose, shape, appearance, and motion features of the moving target. Finally, a graph convolutional network is proposed in~\cite{ferreira2020cag} for generating dance videos from audio information to create natural motions preserving the key movements of different music styles. Nevertheless, these methods focus only on the transfer of a single subject at the same time.

\begin{figure*}[t]
\centering
\includegraphics[width=\textwidth]{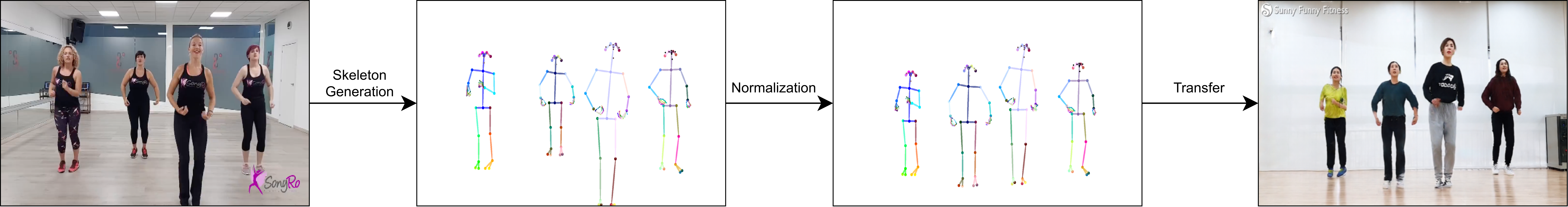}

\caption{An overview of our method. First, a pose detector is used to detect the pose of each actor. For each person, a pose stick figure is generated with a distinct set of colors which is kept consistent over time through pose tracking. Those are then normalized into stick figures for the target domain. Finally, the trained generator is applied.}
\label{fig:norm2}

\end{figure*}
\section{Method}

Starting from a video with a fixed number of source persons and a video with the same number of targets, we aim to generate a new synthetic version of the target video in which the persons now perform the movement seen in the source. Starting from in~\cite{chan2019dance},  the pipeline is divided into three stages – pose detection, global pose normalization, and mapping from normalized skeletons to the target subjects. 

\noindent\textbf{Pose Encoding}~~ Since the focus of our work lies on generalizing pose transfer, we use a pre-trained state of the art model for pose estimation~\cite{cao2017realtime, openpose1, openpose2, openpose3} to produce accurate pose estimation for the input frames. We then generate a colored pose stick figure for each person. In order to reliably learn the appearance of each individual using the poses as intermediate representation, we notice empirically that the stick figures need clear distinct color for each person. If the colors of the body part of two persons are too similar, the model tends to average both appearances and produce less realistic features.

\begin{figure}
	\centering
	\includegraphics[width=0.23\textwidth]{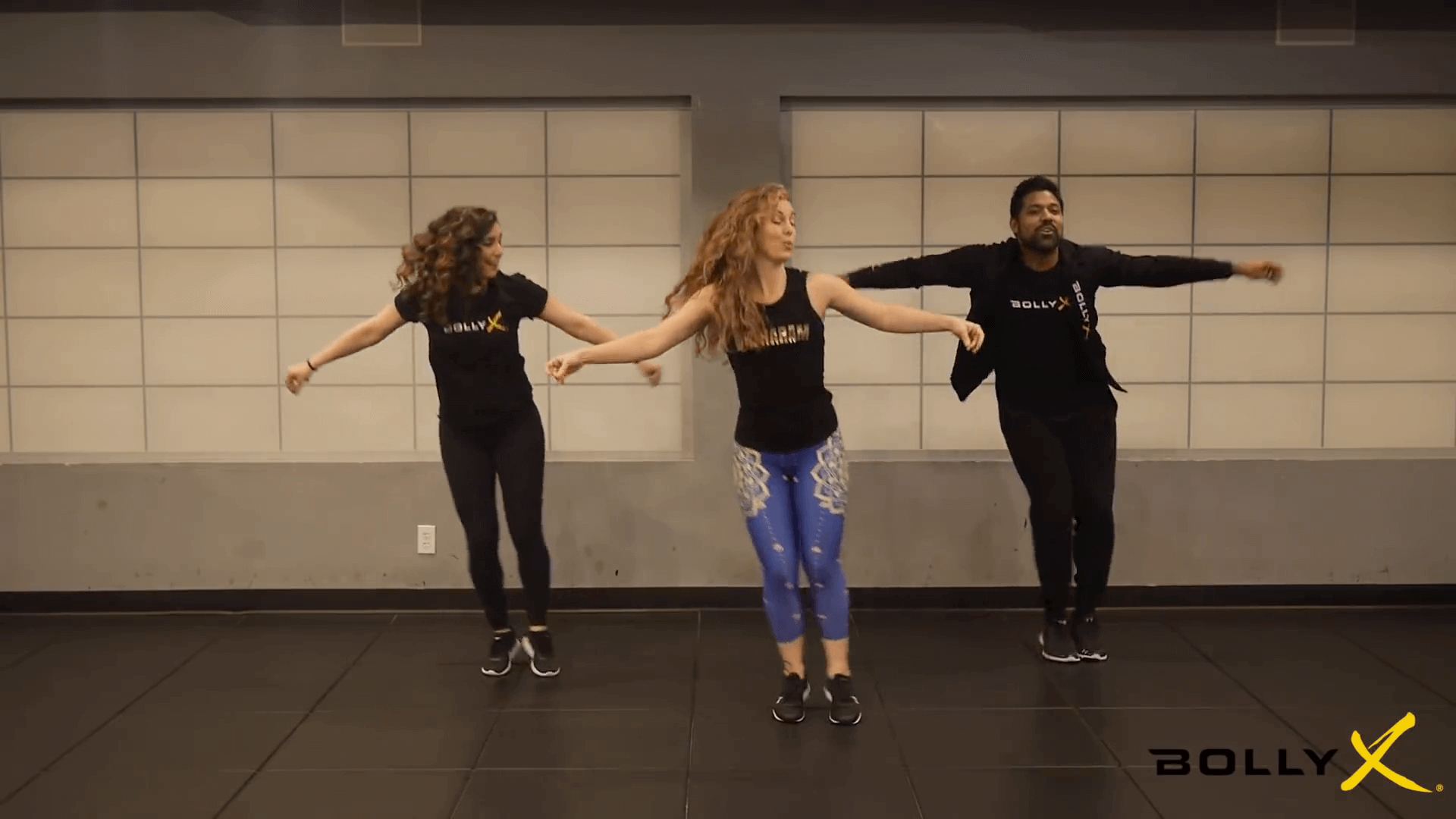} 
    \includegraphics[width=0.23\textwidth]{./images/figure2/source.png} \\
	\includegraphics[width=0.23\textwidth]{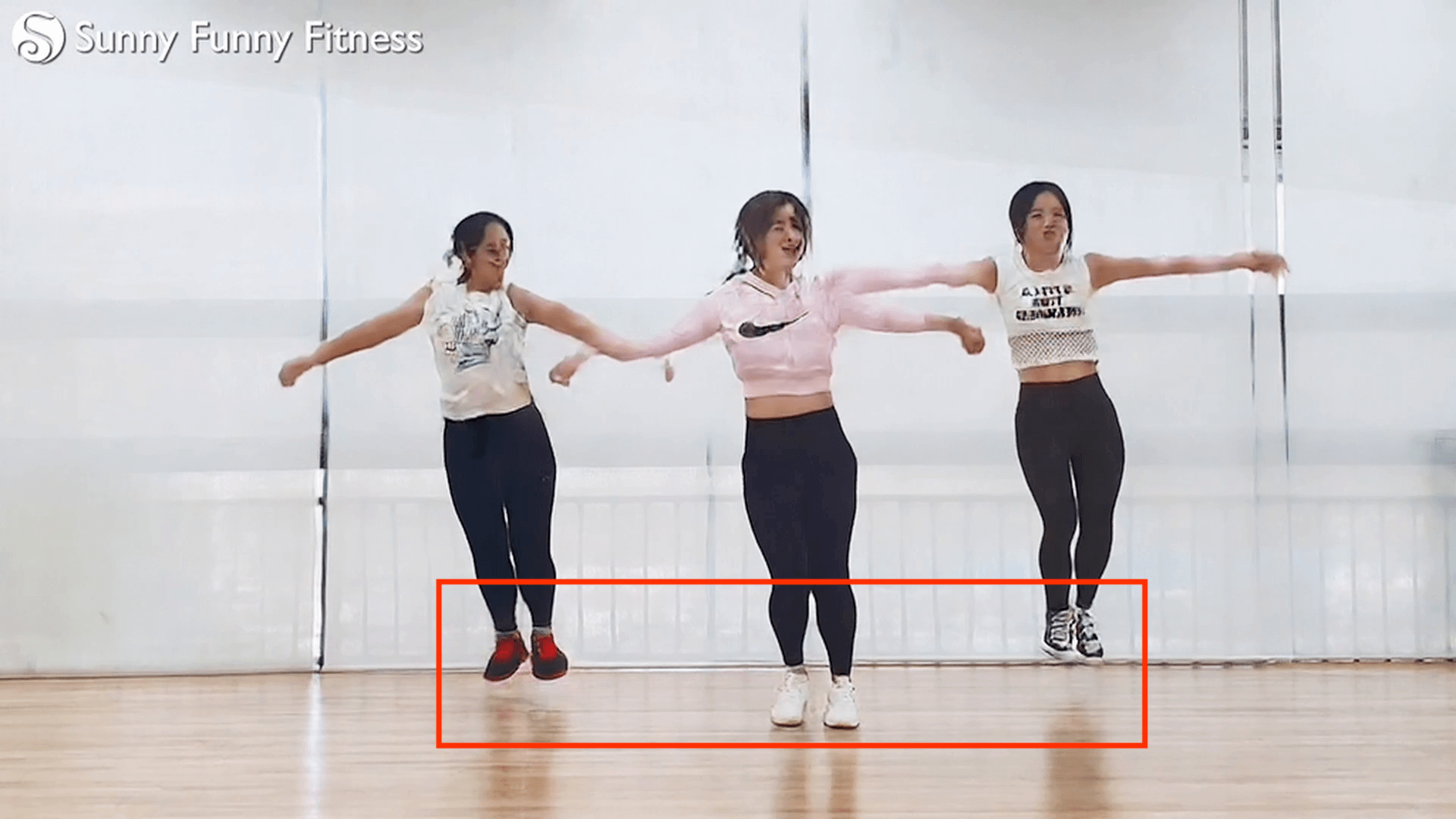}
	\includegraphics[width=0.23\textwidth]{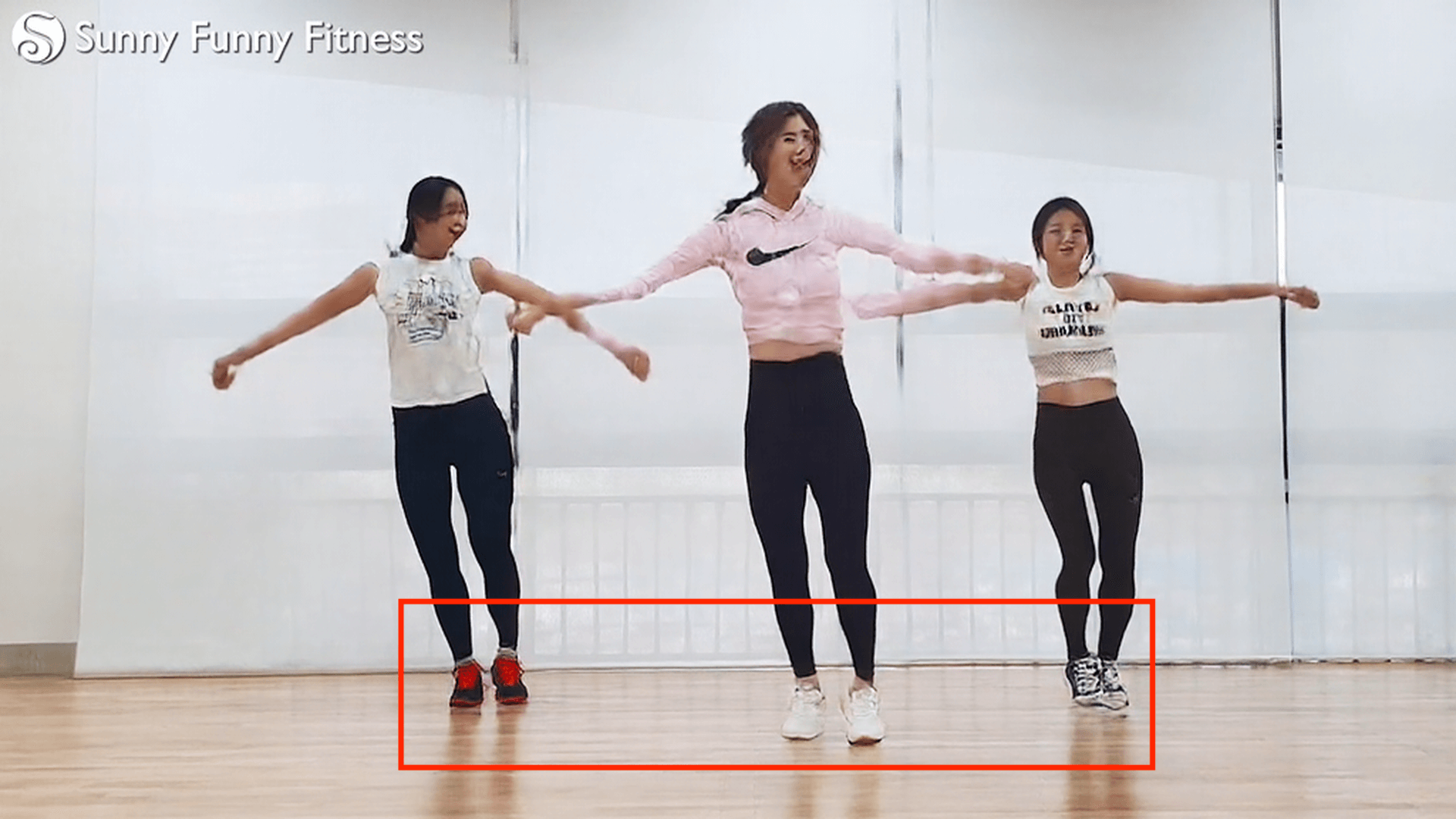}
	\caption{Top: input images. Bottom left: a generated image before normalizing the source keypoints. Bottom right: the same frame after normalizing the source keypoints. In this case the normalization step allows our model to generate the feet of the subjects on the floor.}
	\label{fig:norm1}
\end{figure}

\noindent \textbf{Pose Normalization}~~Since target and source videos have different settings of environment and camera as well as people with different physical appearances such as height and shape, a normalization step between the input and target subjects is needed in order to produce more realistic frames. For instance, in Figure \ref{fig:norm1}, the horizon of the source video is higher than the horizon of the target video which caused the people in the generated video to be above the horizon: their feet are not located on the floor. Moreover, if the person in the source video is taller than the corresponding person in the target video, the subject in the generated frame is abnormally taller. Another instance is if the distance between the people and the camera in the source video is shorter than the distance between the people and the camera in the target video, those in the generated video are peculiarly and disproportionately larger than they should be.

\noindent \textbf{Changes of Place between Source Subjects}~~ If the source subjects change places, there happens to be an identity switch in the target video, meaning an input person will change appearances with another after changing places. To address this, each subject needs to be tracked over time. In this case, poses are tracked before encoding using keypoint correspondences as proposed in~\cite{umer2020self} with scenario-specific adjustments. First, we extend the model for tracking the whole 25 body landmarks available instead of only 17. Although we could also use face and hand landmarks, those aren't predicted as reliably, therefore we decide to use body landmarks only.  Since the number of subjects in each video remains the same, we drop frames where more poses are detected than expected. Typically, in the case where more poses are predicted than there are people in a frame, two poses are assigned to one person which can result in identity switches. For better keypoint heatmap accuracy, we train a keypoint correspondence network with input images of size $512\times512$~pixels instead of $256\times256$~pixels. As the persons are a closed set no similarity threshold is used and poses are always assigned in a greedy fashion to the ids of the closed set.

\noindent \textbf{Pose to Video Translation}~~ After preprocessing the tracked skeletons are then used as input to our model. An overview of our method is given in Figure~\ref{fig:norm2}. Our model is based on an adversarial conditional GAN setup, trained in three stages: global, local and faces. For more details, we refer to~\cite{chan2019dance}.

\section{Experiments}
\subsection{Setup}
A separate model is trained on collected frames for each training video of 2, 3, 4 and 5 people videos at $1024\times512$ respectively. This is followed by an evaluation using unseen test frames in order to evaluate the efficiency of our approach.
Each model is trained separately in three stages. In the first stage, a global generator is used for training the model. In the second stage, the model is refined with a local enhancer generator and finally FaceGAN is used in the last stage.
For our experiments we use the Adam optimizer with learning $rate = 0.0002$ and $B = 0.999$. For all experiments the batch size is set to $1$. Also we set $\lambda_{VGG}=10$. We trained our models for about $168,000$ iterations which required totally about $35$ hours on a RTX 2080Ti. As baseline we reproduce the results from~\cite{chan2019dance} for single to single person motion transfer.

\noindent \textbf{Evaluation Metrics}
We measure the quality of the synthesized frames using four metrics: 1) Peak Signal-to-Noise Ratio (PSNR)  measures the similarity of the pixel-level images between generated images, 2) Structural Similarity (SSIM)~\cite{ssim} between two images by comparing three factors of luminance, contrast, and structure, 3) Learned Perceptual Image Patch Similarity (LPIPS)~\cite{lpips} to measure the perceptual similarity between synthesized images and ground truth, and 4) Frechet Inception Distance (FID)~\cite{fid} to measure the quality of frames of generated videos. We strive for high metrics for 1) and 2) and for smaller metrics for 3) and 4).

\subsection{Quantitative Results}
We compare our approach quantitatively for a different number of people. We perform the evaluation on a held-out test data and report our results in Table~\ref{tab:qualitative_eval_no-face}. We first report results for single-to-single transfer as a baseline. For this model almost four times more data is available for training than for multiple person transfer, which may partially explain the performance gap between the baseline and our models. Overall our models provide satisfying results for the few constraints we chose to apply for collecting the video pairs. Surprisingly our 5-to-5 model performs especially well by FID and LPIPS metrics, which means the results should be more convincing in the human eye. In order to refine our results and since acceptable face keypoints for the 3-to-3 and 5-to-5 videos were available, we added 60 supplemental face landmarks to our model and report our findings in Table~\ref{tab:qualitative_eval_with-face}. This addition brought a strong improvement in terms of FID to the 3-to-3 model and a smaller improvement to the 5-to-5 model. However, we emphasize that these improvements are largely reliant on the quality of the pose estimator's predictions. Therefore, such additional facial landmarks may not always be available or of sufficient quality.

\begin{table}
	\centering
	\begin{tabular}{l c c c c c c} \toprule
\textbf{Model} & \textbf{FID$\downarrow$}& \textbf{LPIPS$\downarrow$}& \textbf{PSNR$\uparrow$}  & \textbf{SSIM$\uparrow$}  \\ \midrule
        1-to-1             & $20.843$  & $0.060$ & $37.837$ & $0.950$ \\\midrule
		2-to-2           & $25.280$  & $0.085$ & $36.576$ & $0.947$ \\\midrule %Visiosys
		3-to-3                & $24.364$      & $0.178$ & $34.135$    & $0.865$\\\midrule
		4-to-4              & $26.797$   & $0.211$ & $33.319$     & $0.843$ \\\midrule
		5-to-5             & $8.510$ & $0.088$  & $35.666$      & $0.925$ \\\bottomrule
		
	\end{tabular}
	\caption{1 Person with 23,000 Frames as in~\cite{chan2019dance}. The other models are trained with around 5,600 frames due to unavailability of more frames.}
	\label{tab:qualitative_eval_no-face}
\end{table}

\begin{table}
	\centering
	\begin{tabular}{l c c c c c c c} \toprule
		
\textbf{Model} & \textbf{F}  & \textbf{FID$\downarrow$} & \textbf{LPIPS$\downarrow$} & \textbf{PSNR$\uparrow$}  & \textbf{SSIM$\uparrow$} \\ \midrule
		3-to-3          &      & $24.364$      & $0.178$ & $34.135$    & $0.865$\\\midrule
		5-to-5         &    & $8.510$ & $0.088$  & $35.666$      & $0.925$ \\ \hline\hline 
		3-to-3        &    \checkmark    & $20.210$       & $0.173$  & $34.214$     &$0.865$  \\\midrule
		5-to-5        & \checkmark    & $8.086$  & $0.088$  & $33.110$       & $0.830$
 \\\bottomrule
	\end{tabular}
	\caption{Our best models are further optimized using 68 face keypoints instead of only eight.}
	\label{tab:qualitative_eval_with-face}
\end{table}

\subsection{Qualitative Results}
\begin{figure*}
\centering
\subfloat{

	\includegraphics[width=0.32\textwidth]{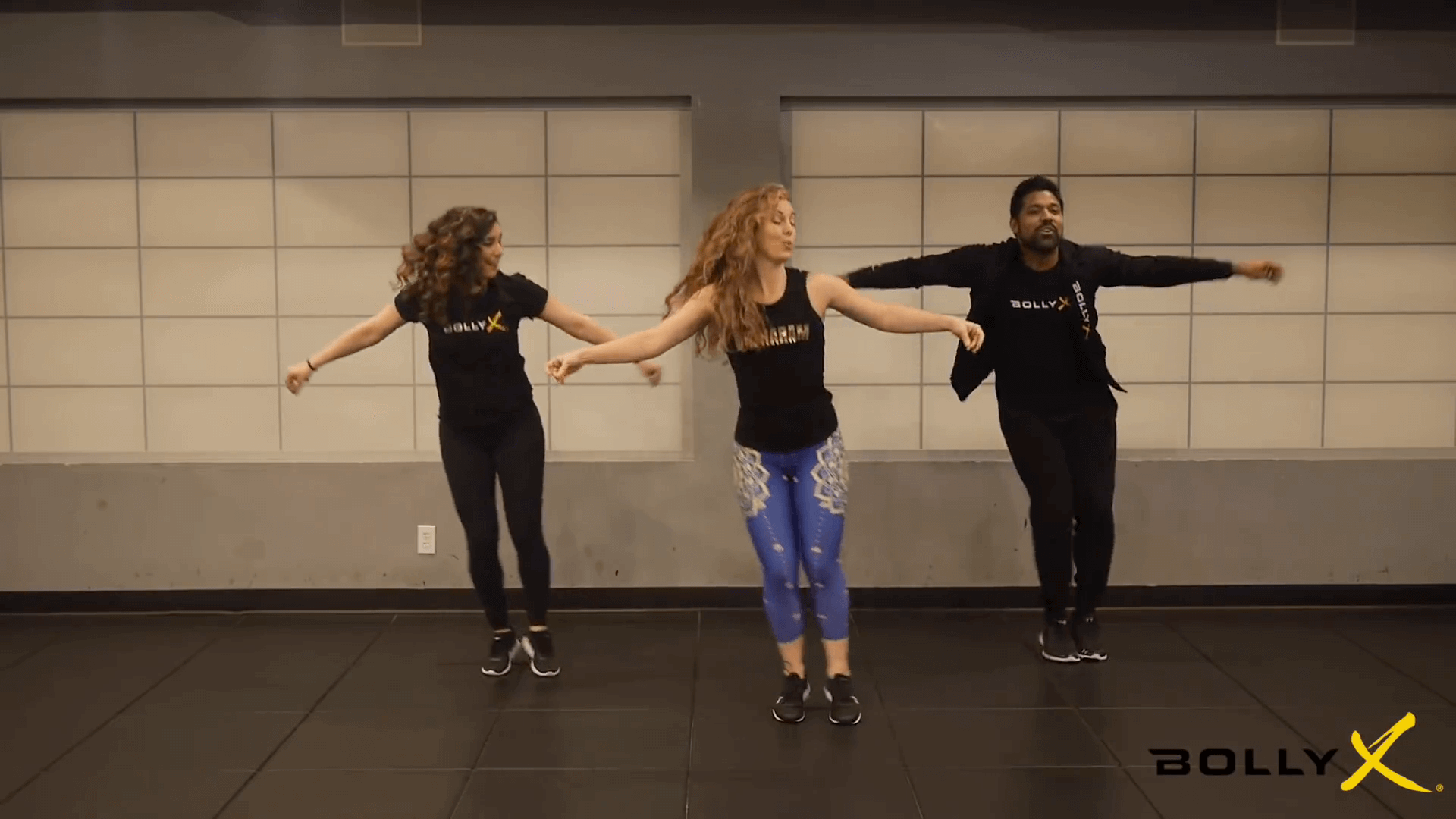}
	\includegraphics[width=0.32\textwidth]{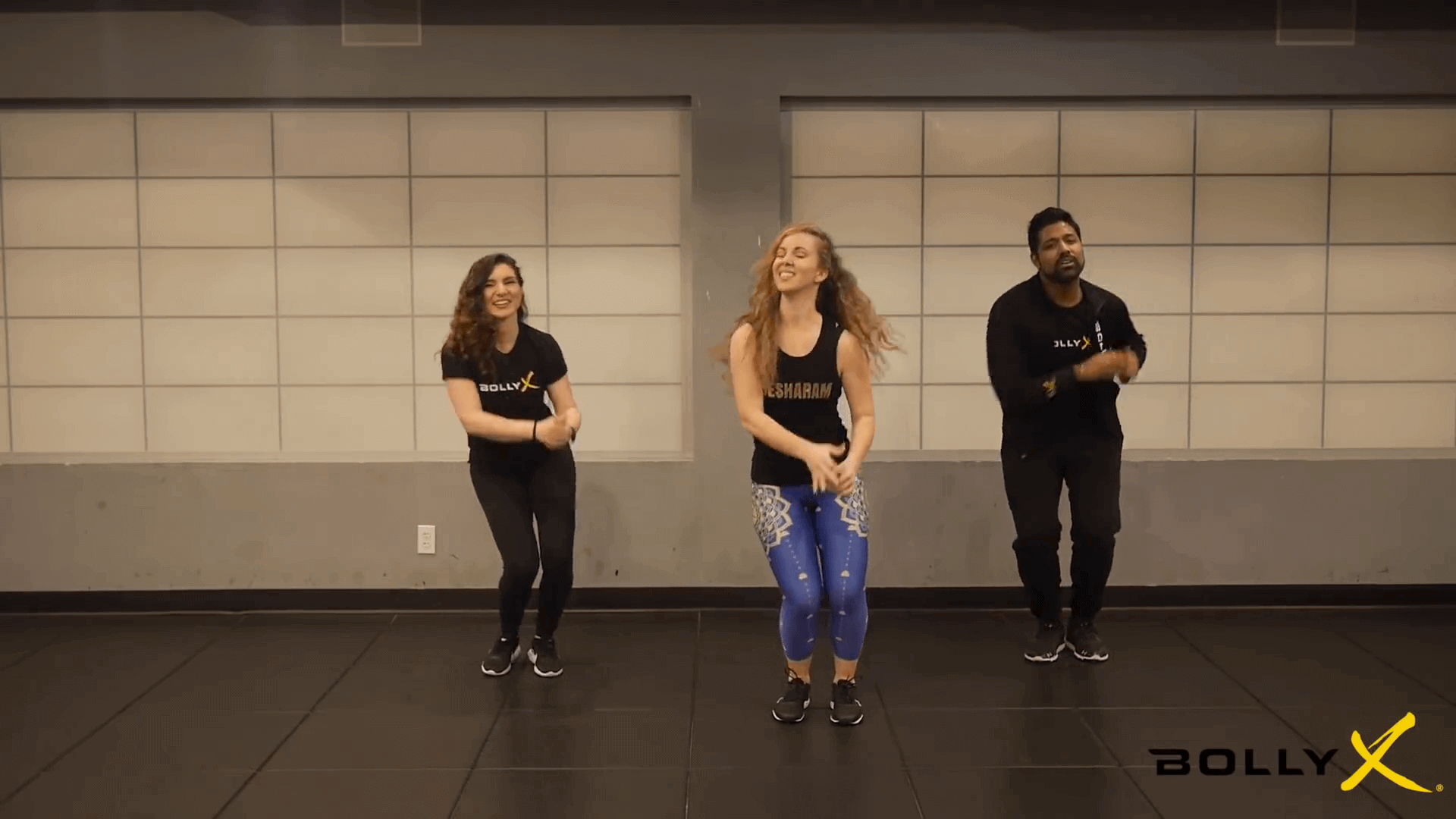}
	\includegraphics[width=0.32\textwidth]{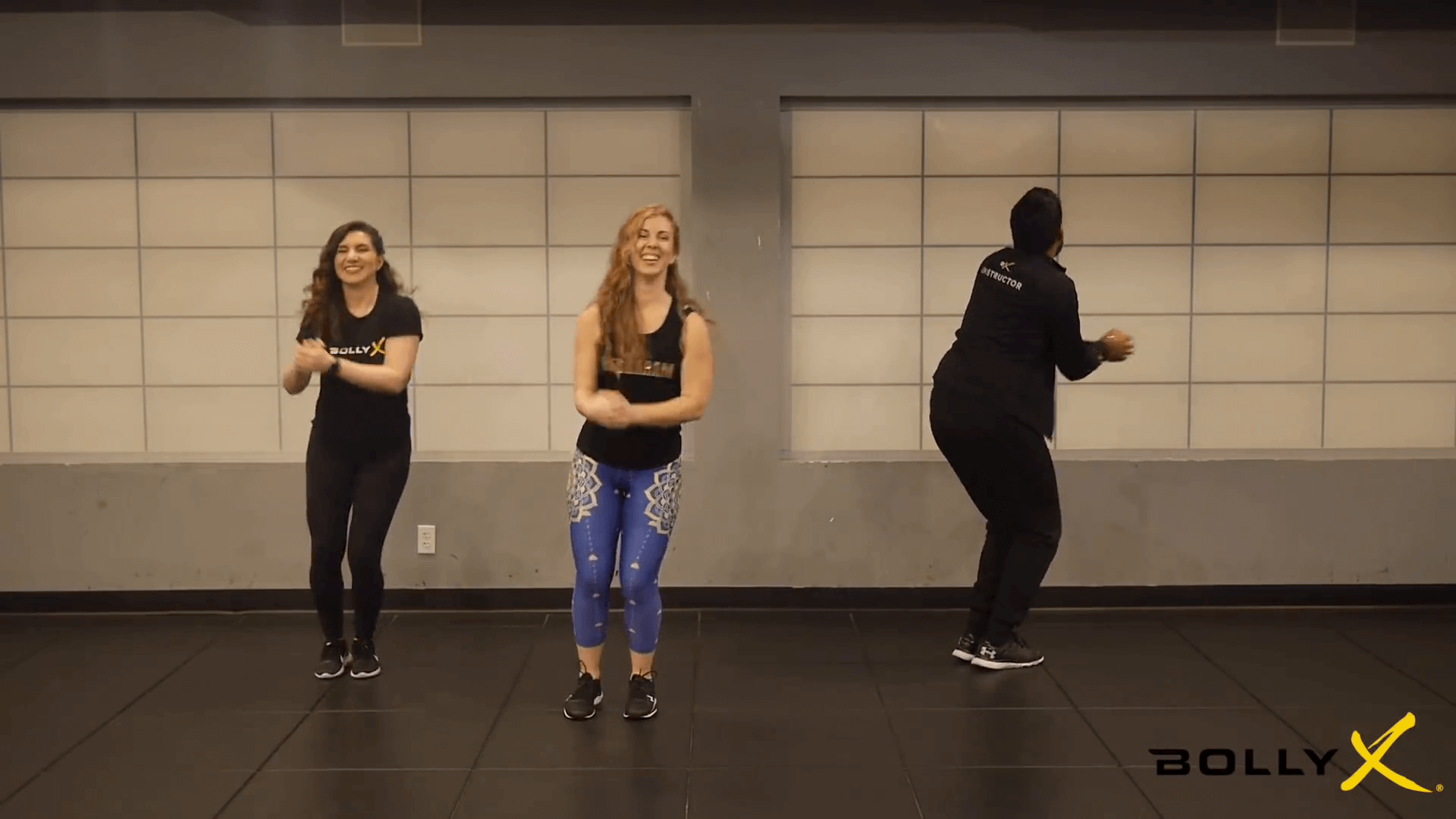}

} \qquad
\subfloat{

	\includegraphics[width=0.32\textwidth]{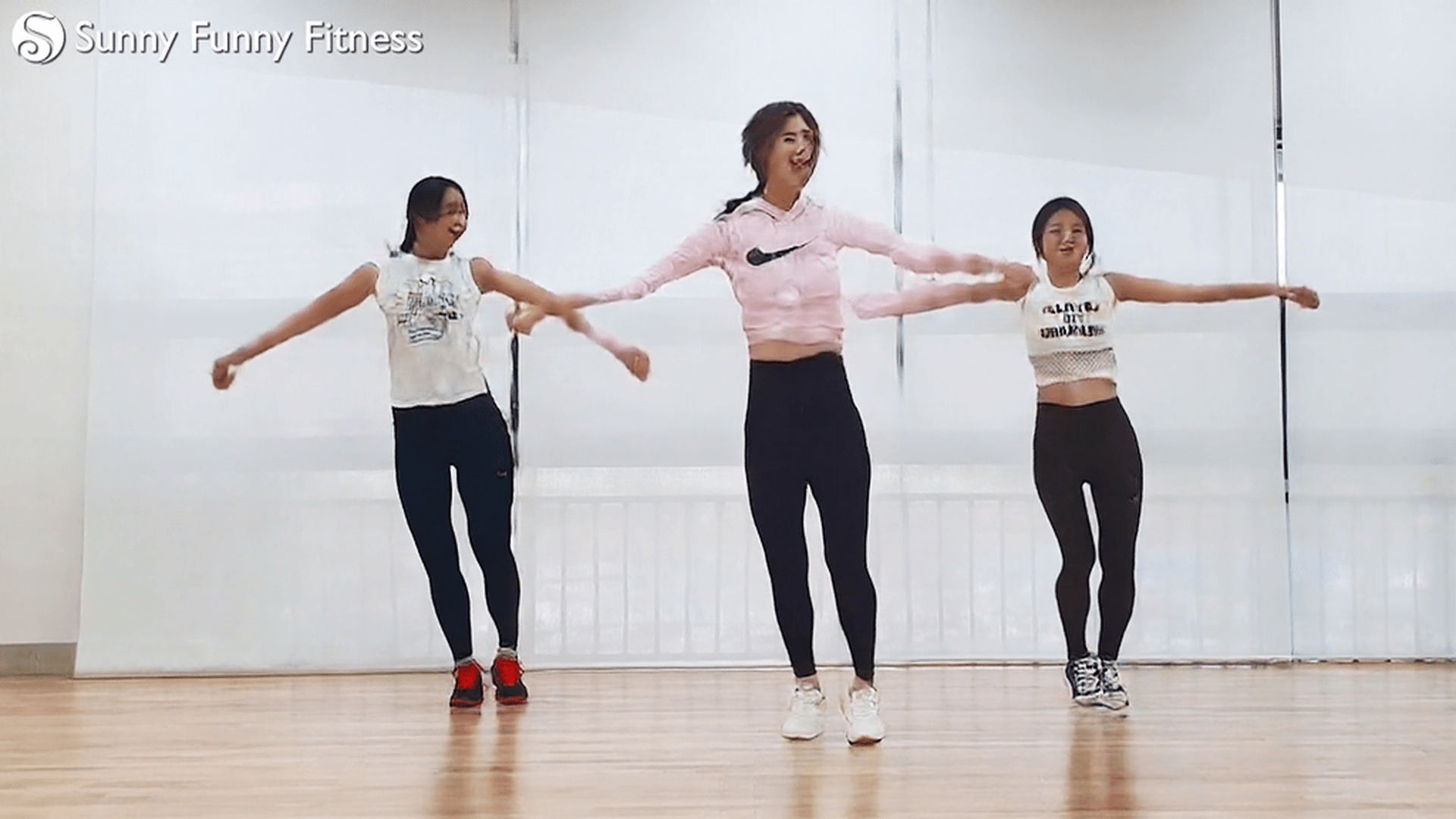}
	\includegraphics[width=0.32\textwidth]{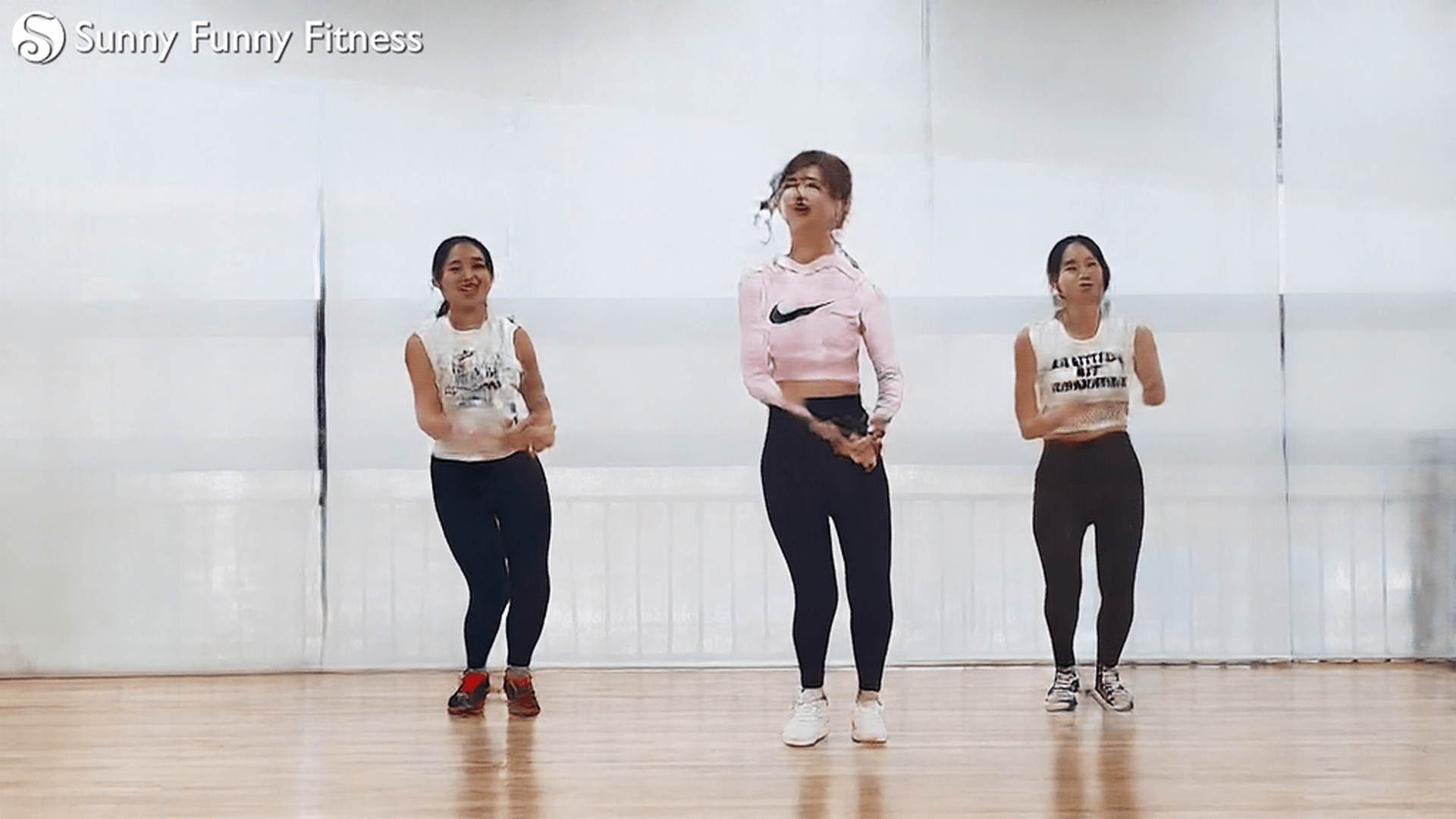}
	\includegraphics[width=0.32\textwidth]{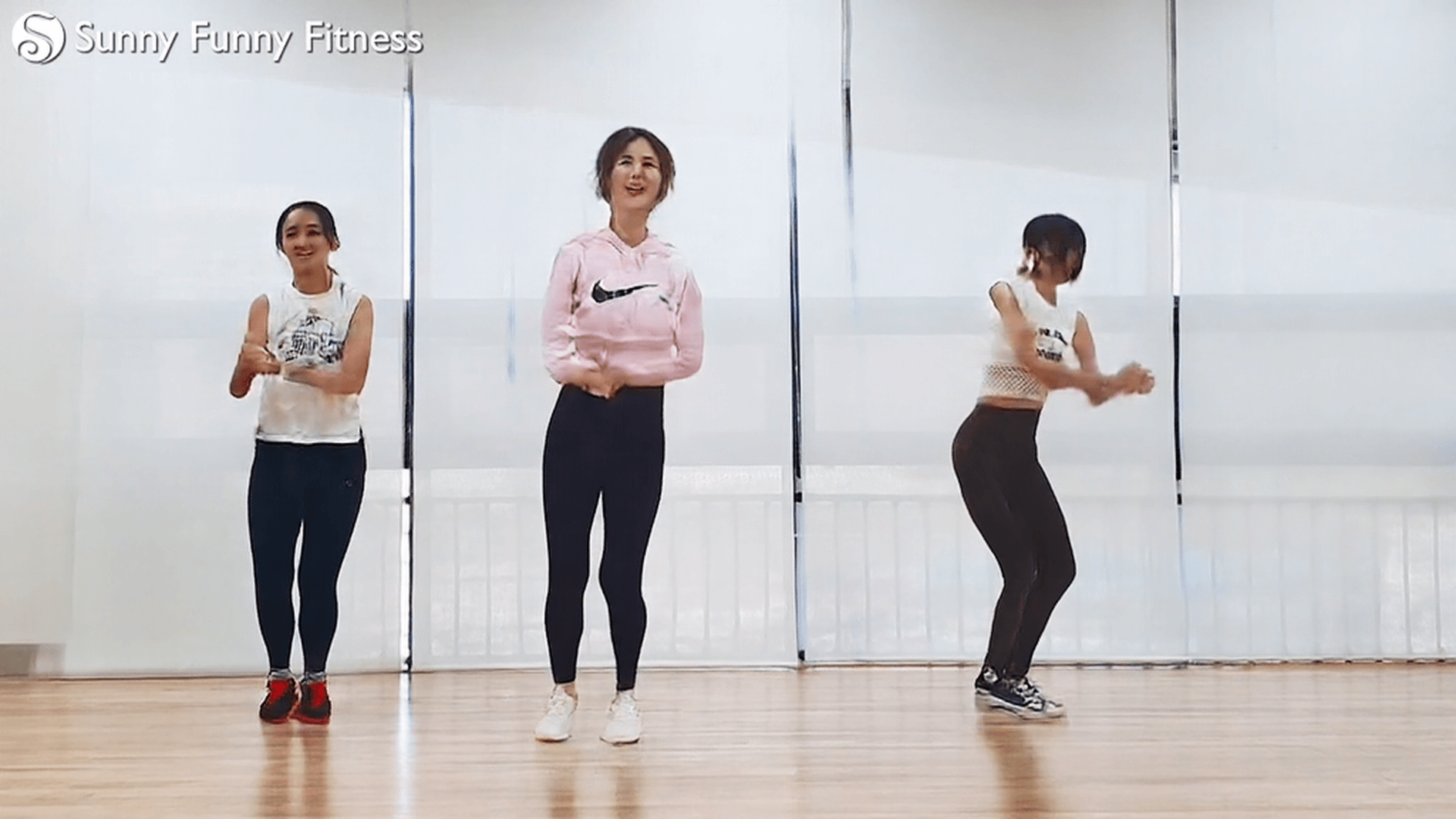}
} \qquad
\subfloat{

	\includegraphics[width=0.32\textwidth]{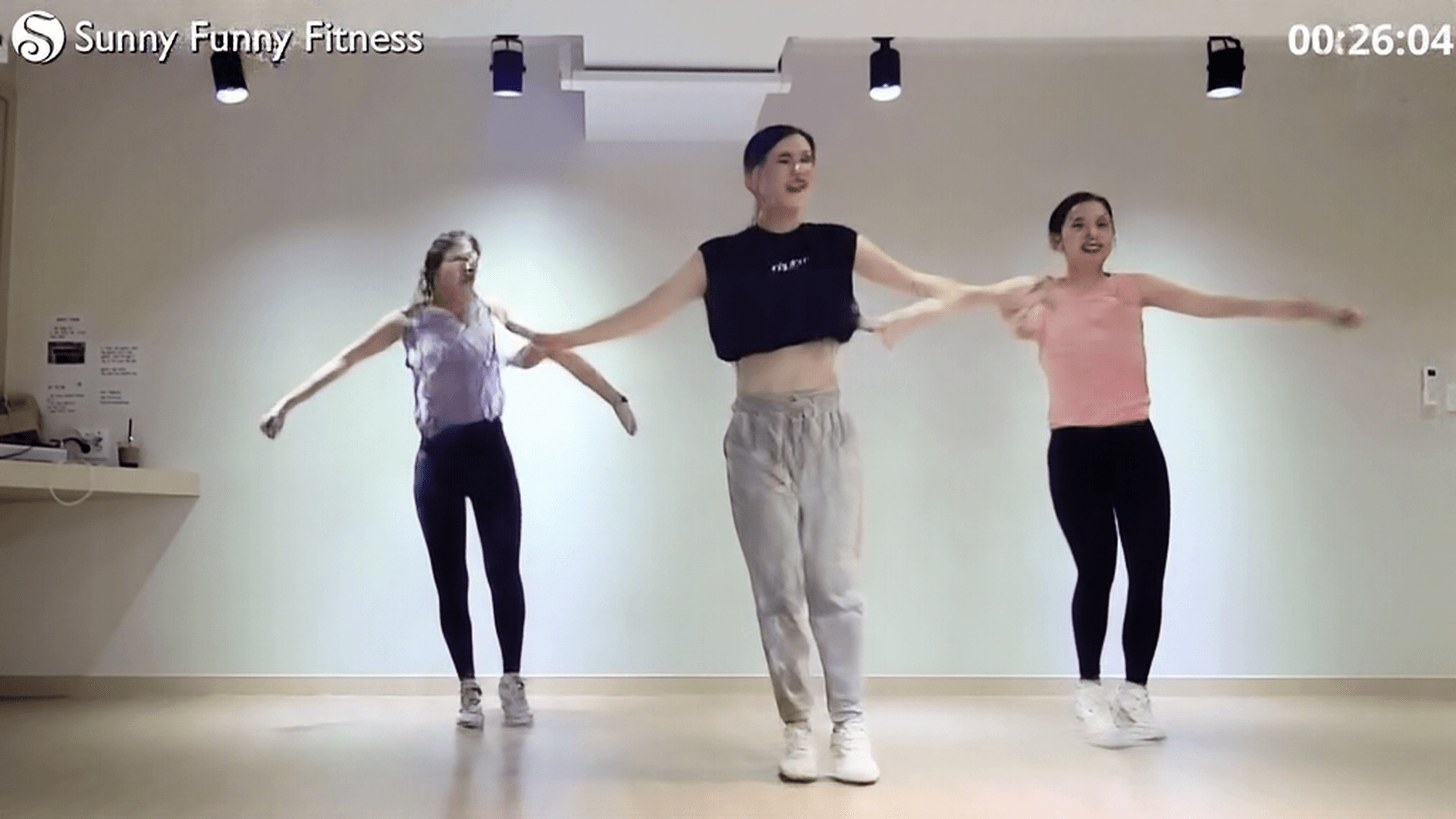}
	\includegraphics[width=0.32\textwidth]{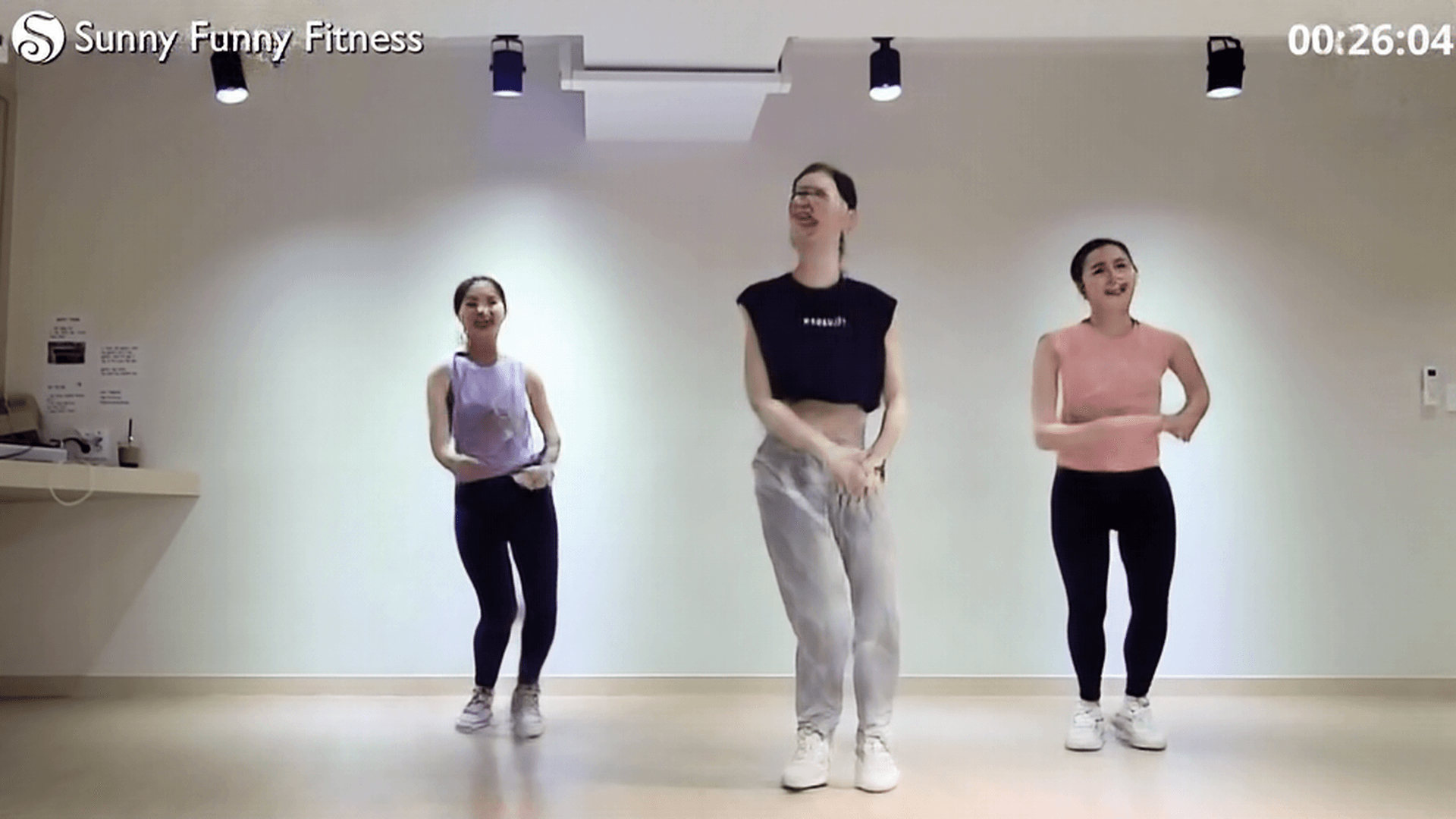}
	\includegraphics[width=0.32\textwidth]{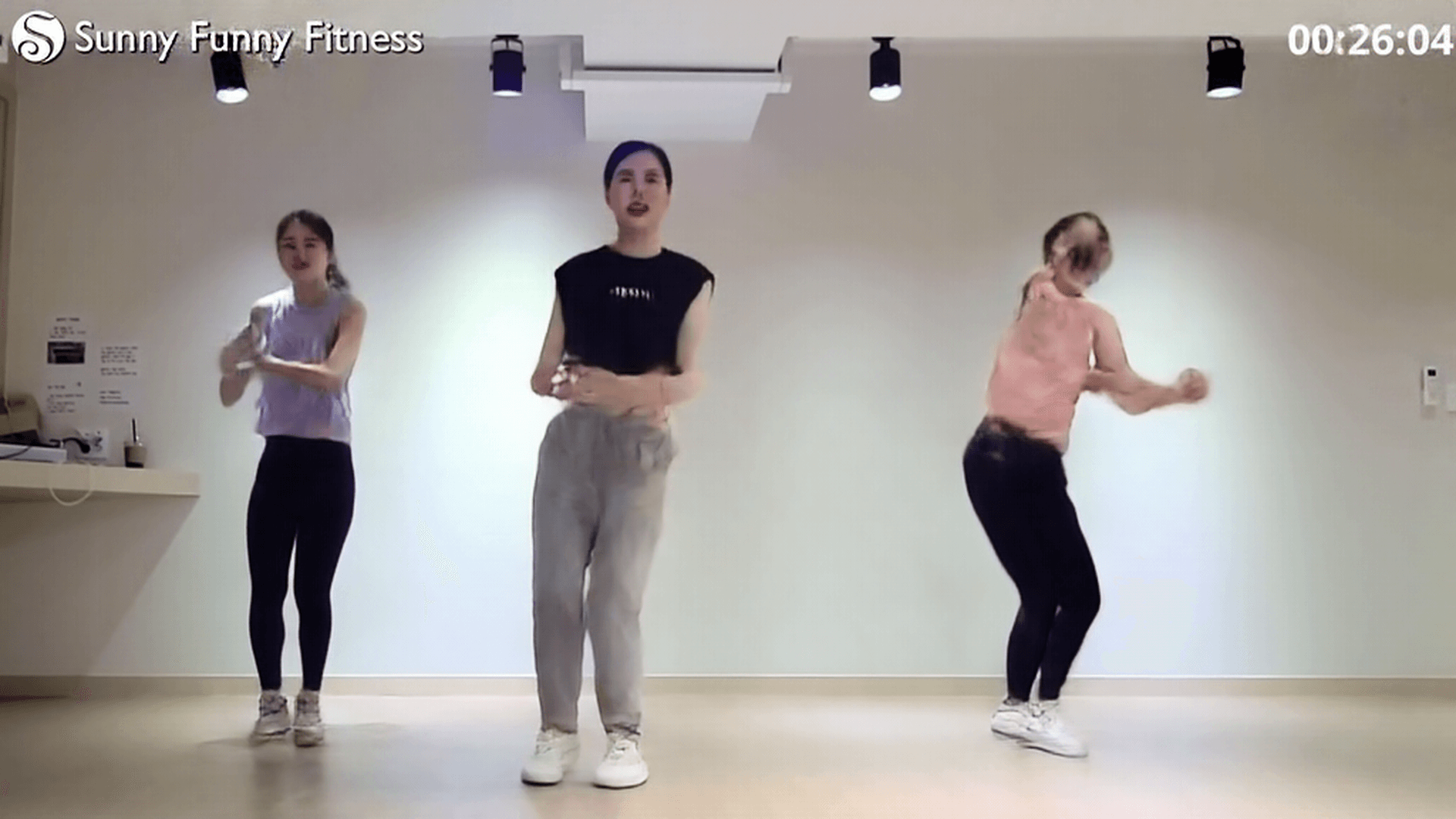}

}
\caption{Given a source video~\cite{3person1} (top) and two different target videos from~\cite{3person_sunny1} (middle) and ~\cite{3person_sunny2} (bottom). Our approach transfers the movements from the people in (top) to the people in middle and last row. While the middle row handles the turning face (right), the last row can't manage the face pose and produces a dark artifact in the center of the face.}
\label{fig:transfer_2}
\end{figure*}

\begin{figure*}
\centering
\subfloat{

	\includegraphics[width=0.32\textwidth]{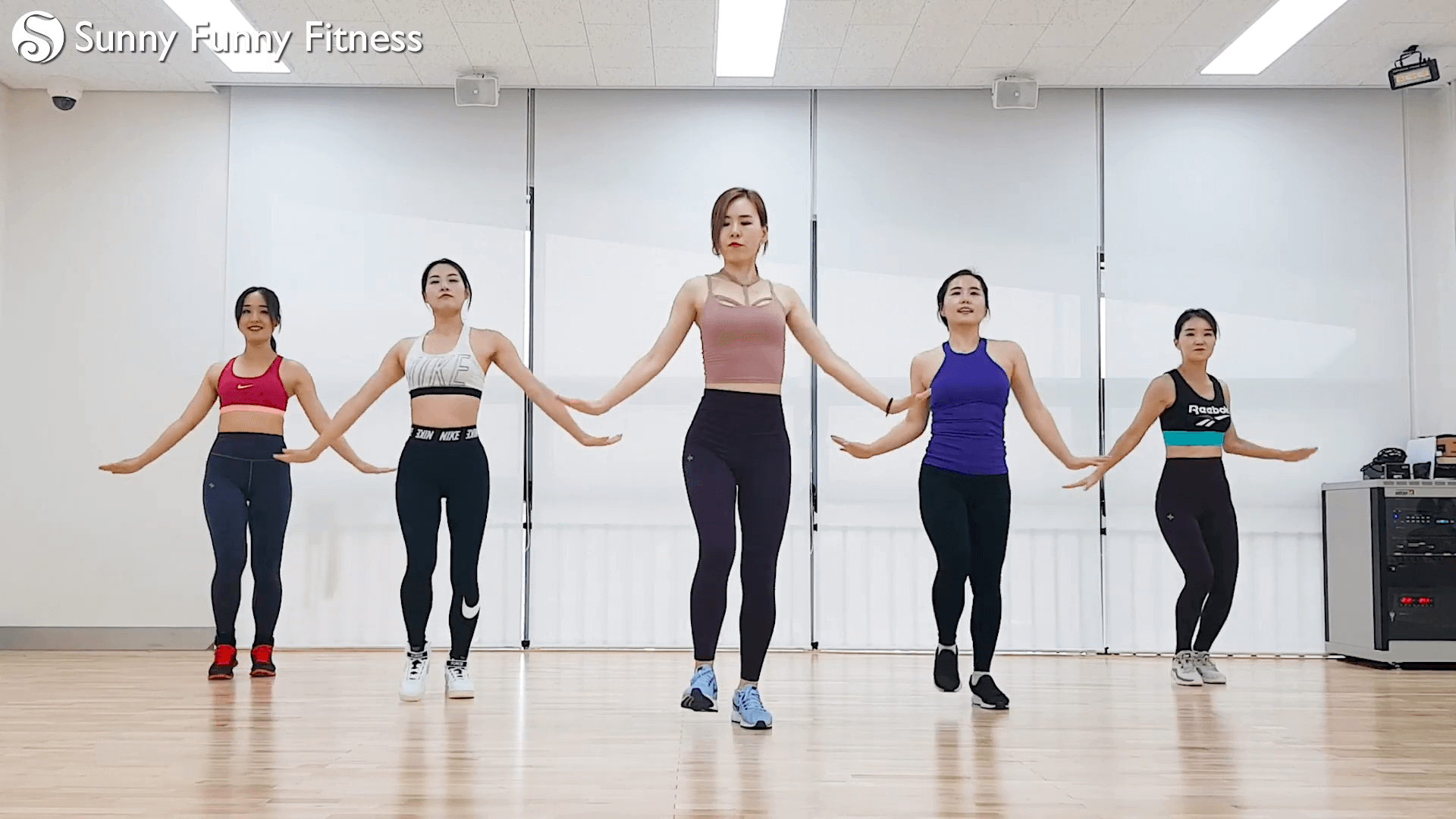}
	\includegraphics[width=0.32\textwidth]{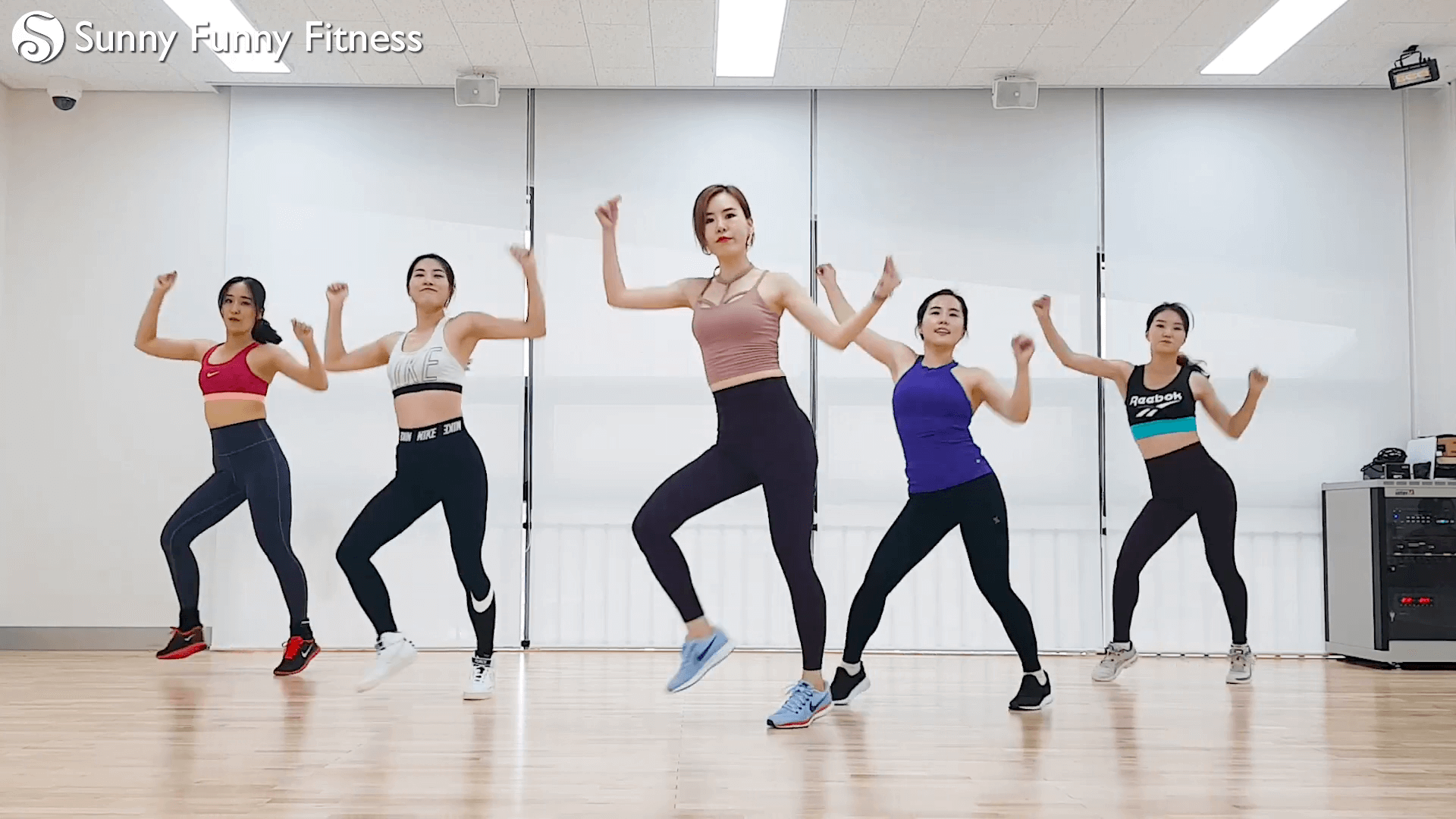}
	\includegraphics[width=0.32\textwidth]{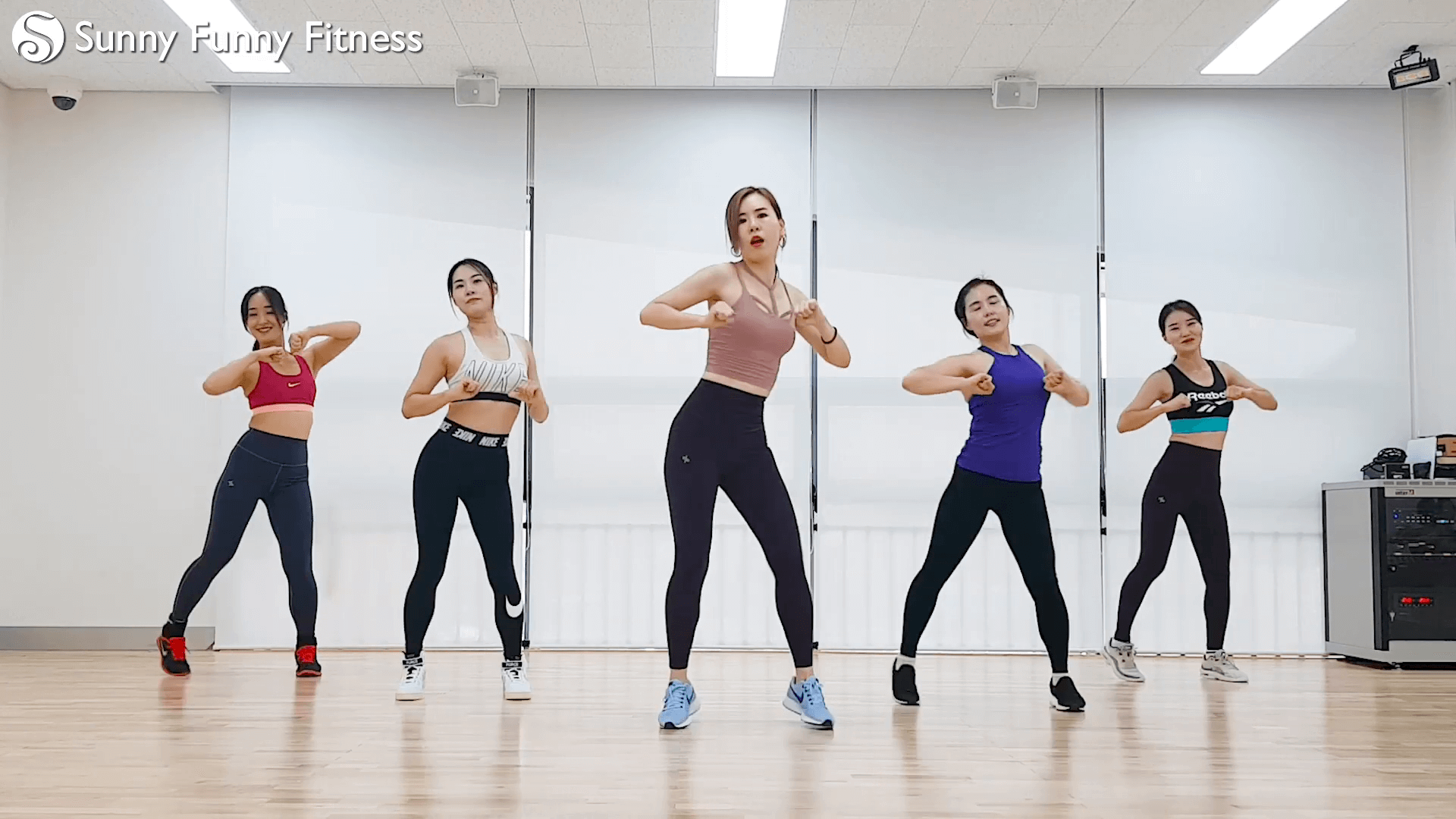}
	
} \qquad
\subfloat{

	\includegraphics[width=0.32\textwidth]{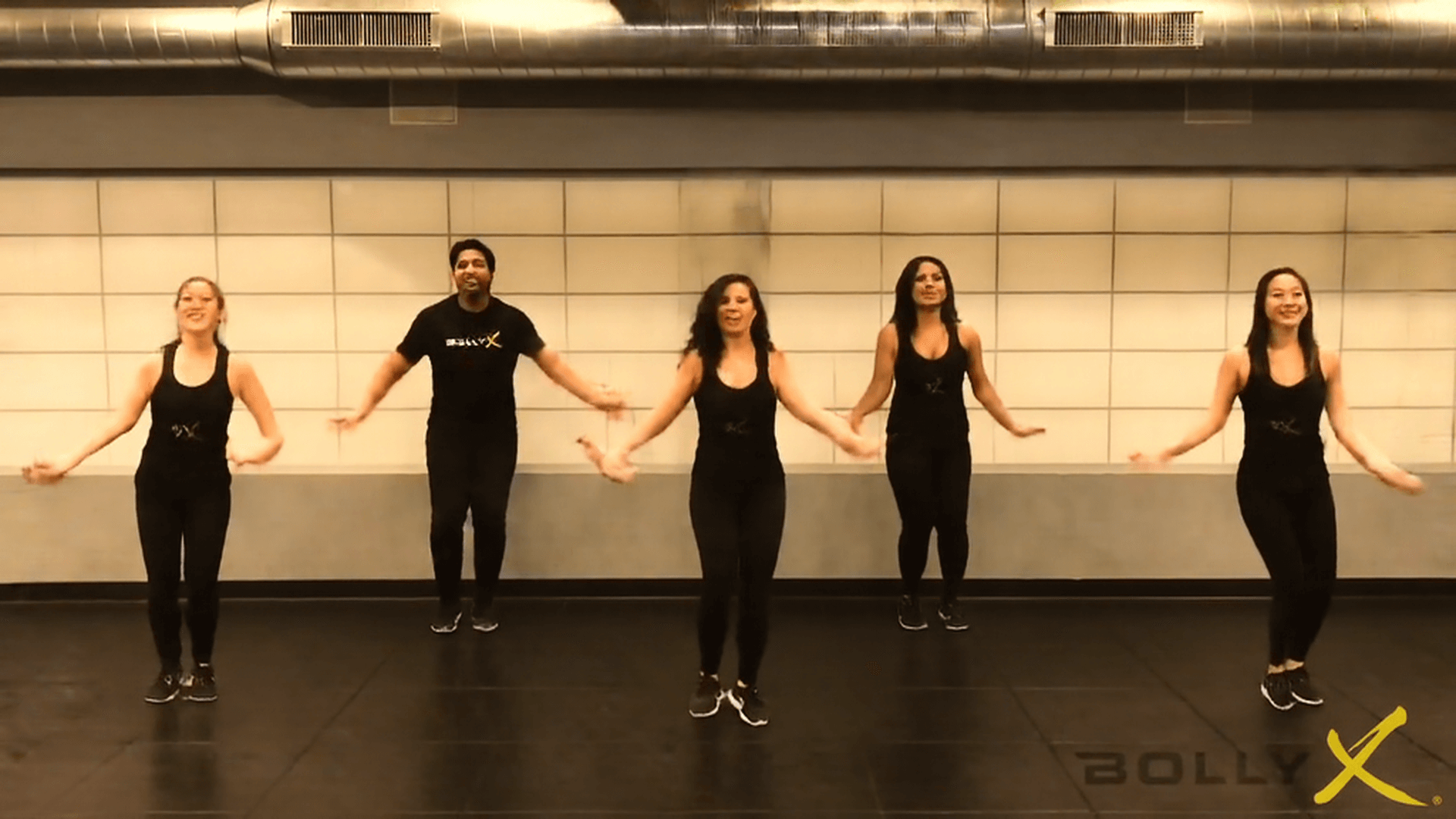}
	\includegraphics[width=0.32\textwidth]{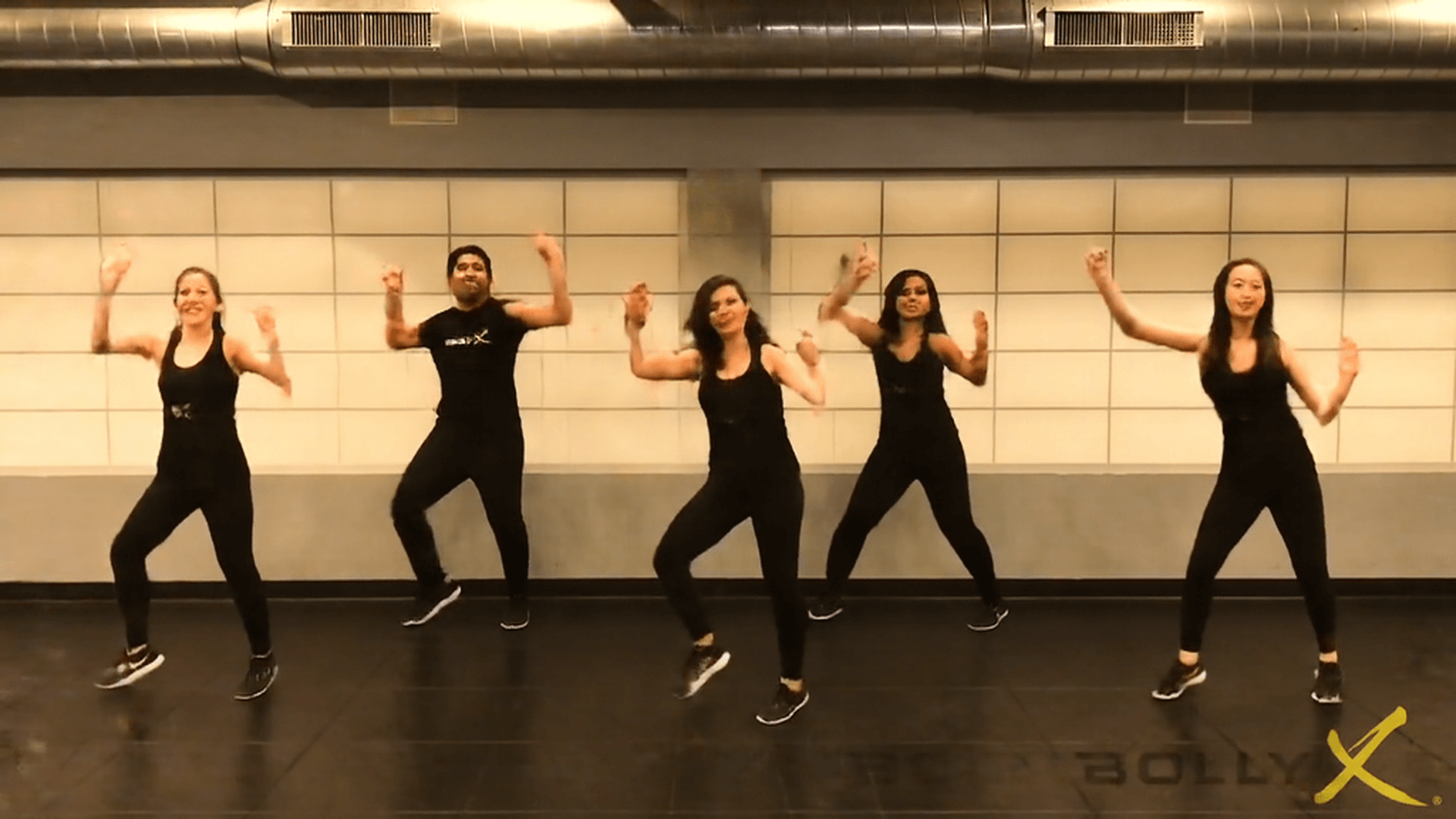}
	\includegraphics[width=0.32\textwidth]{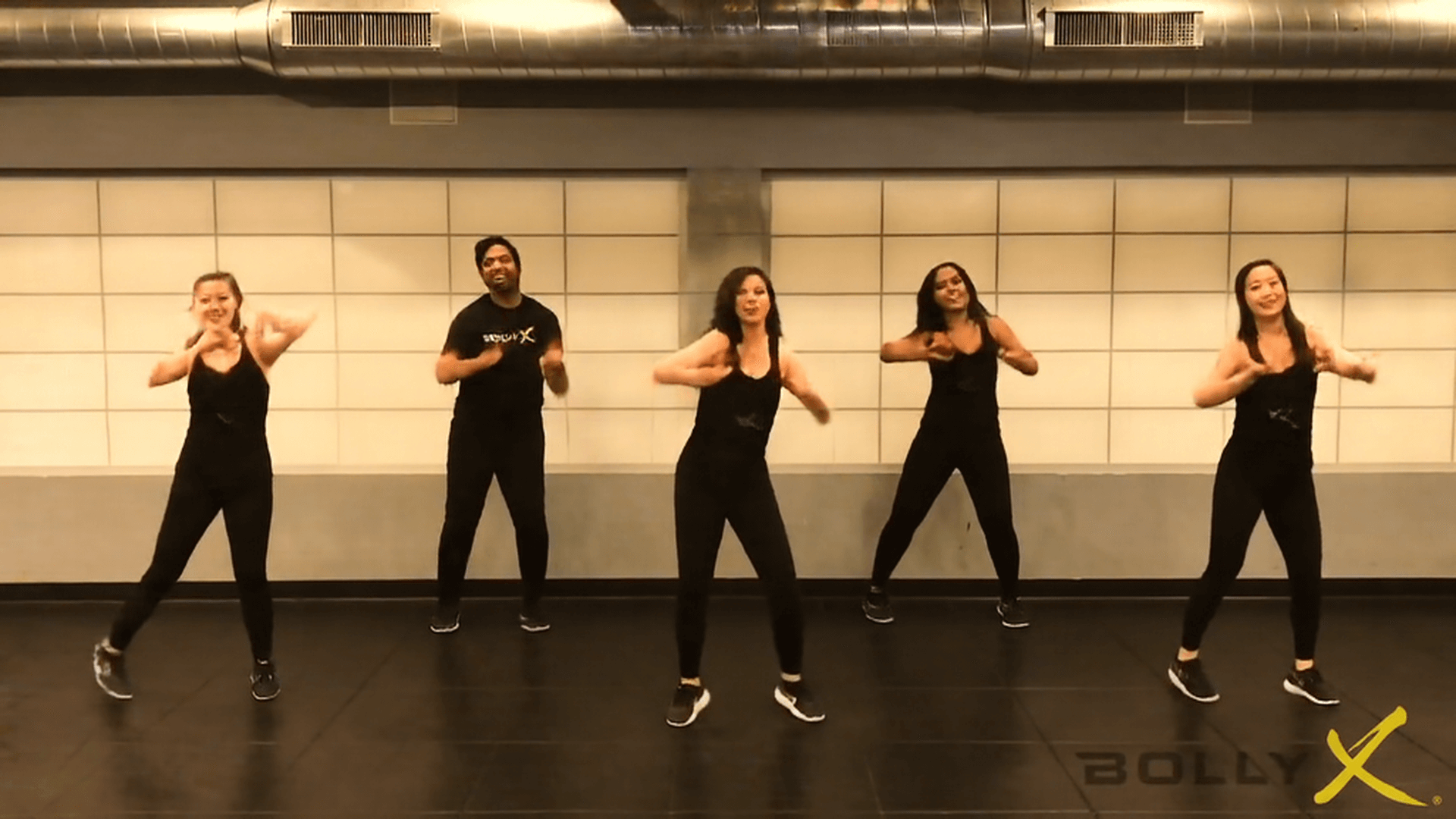}

}
    \caption{Results for our 5-to-5 person model. Input appearance from~\cite{5person1} (top) followed by our results (bottom) on~\cite{5person2}. Due to relatively uniform target clothes, the results are particularly smooth.}
    \label{fig:results:5p_good}
\end{figure*}

\begin{figure*}
\centering
\subfloat{

	\includegraphics[width=0.32\textwidth]{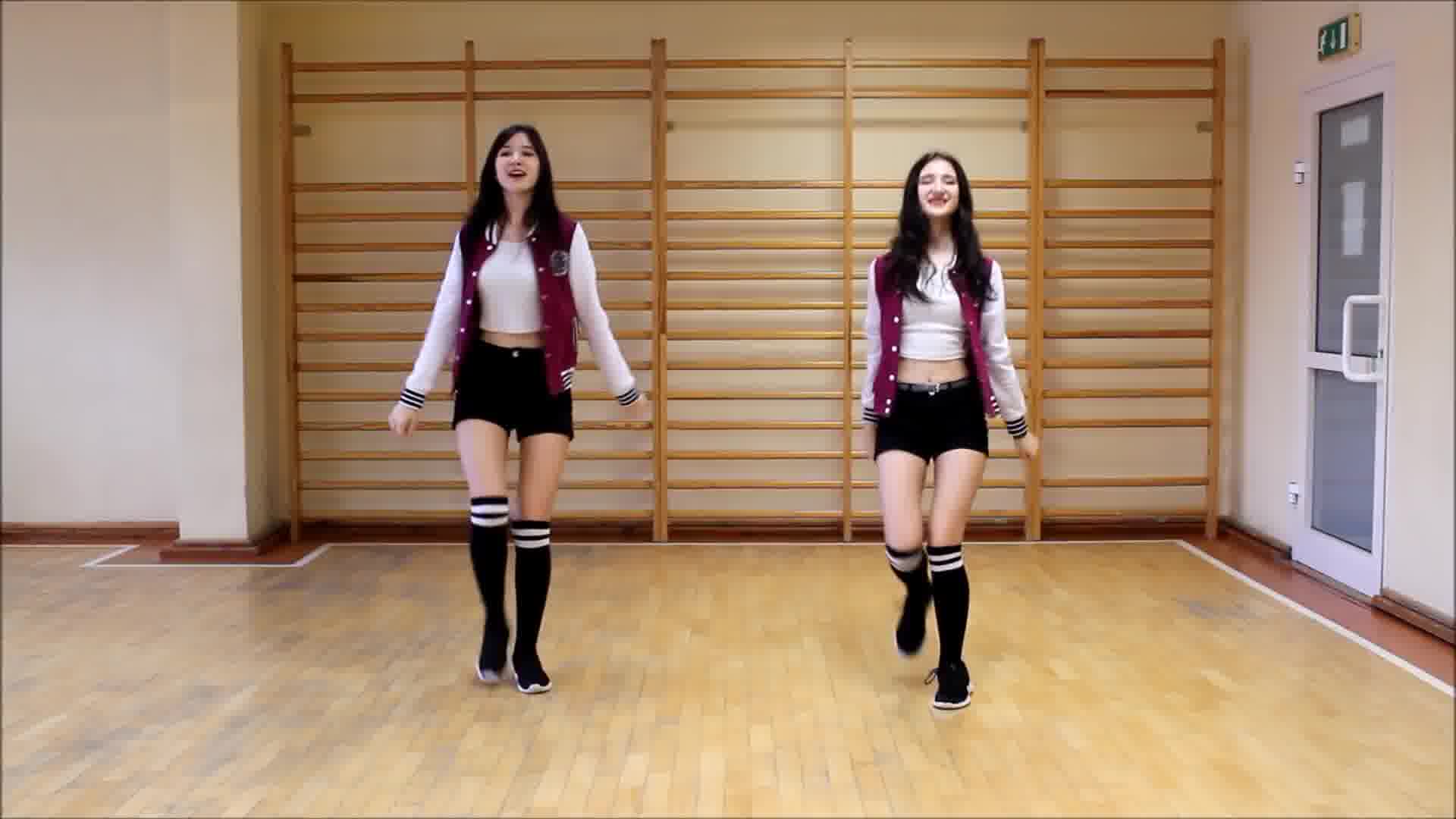}
	\includegraphics[width=0.32\textwidth]{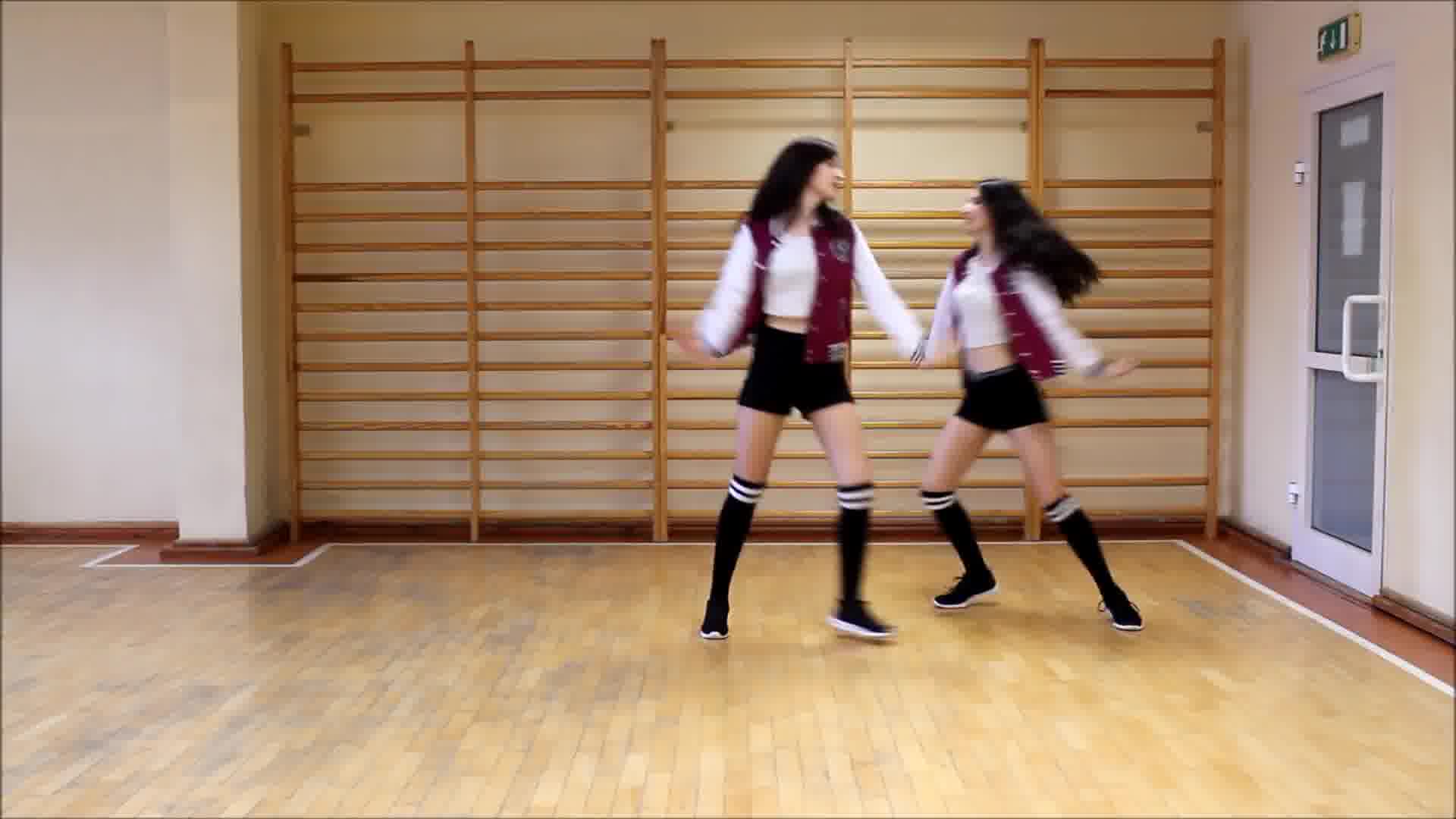}
	\includegraphics[width=0.32\textwidth]{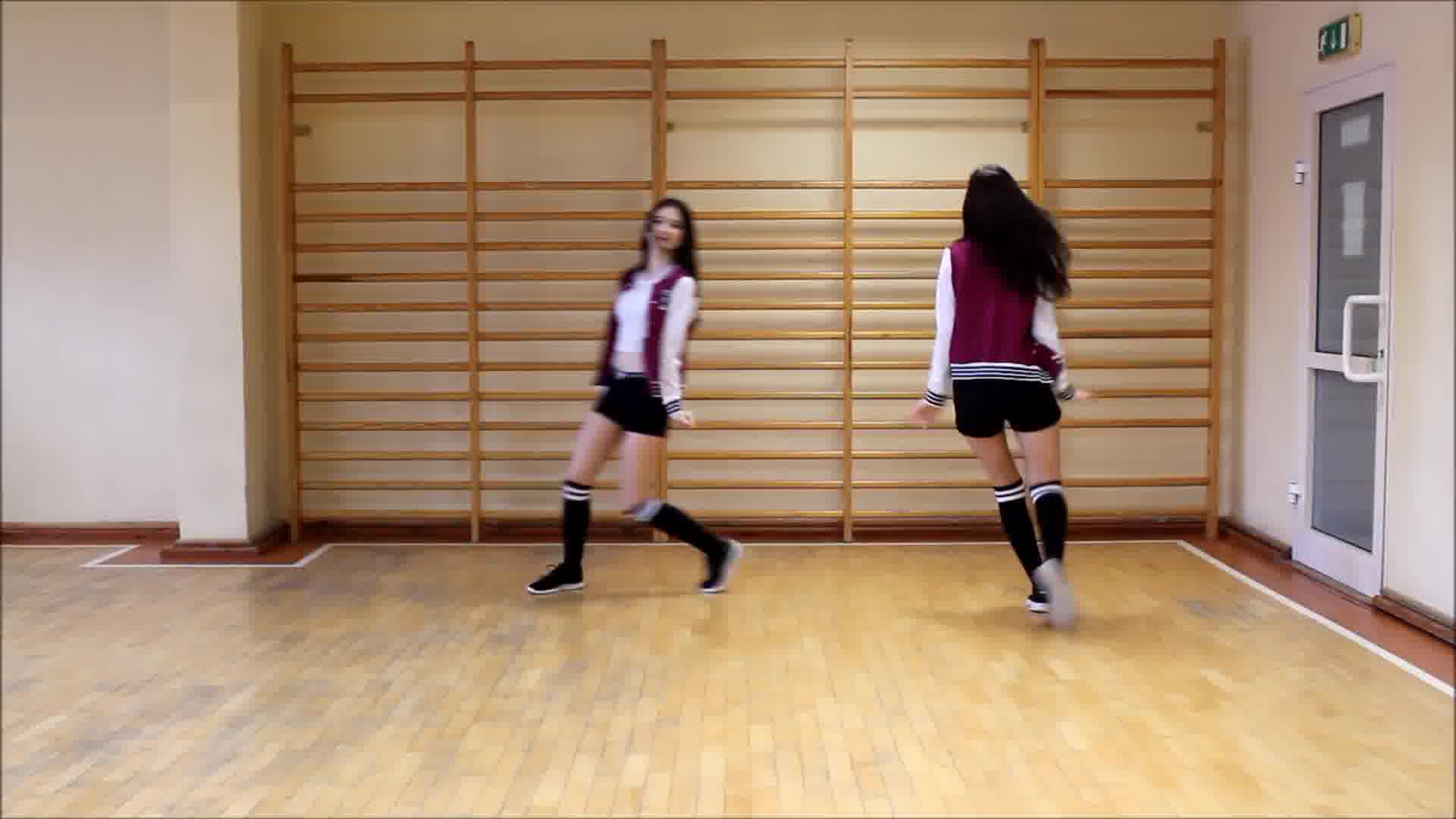}

} \qquad
\subfloat{

	\includegraphics[width=0.32\textwidth]{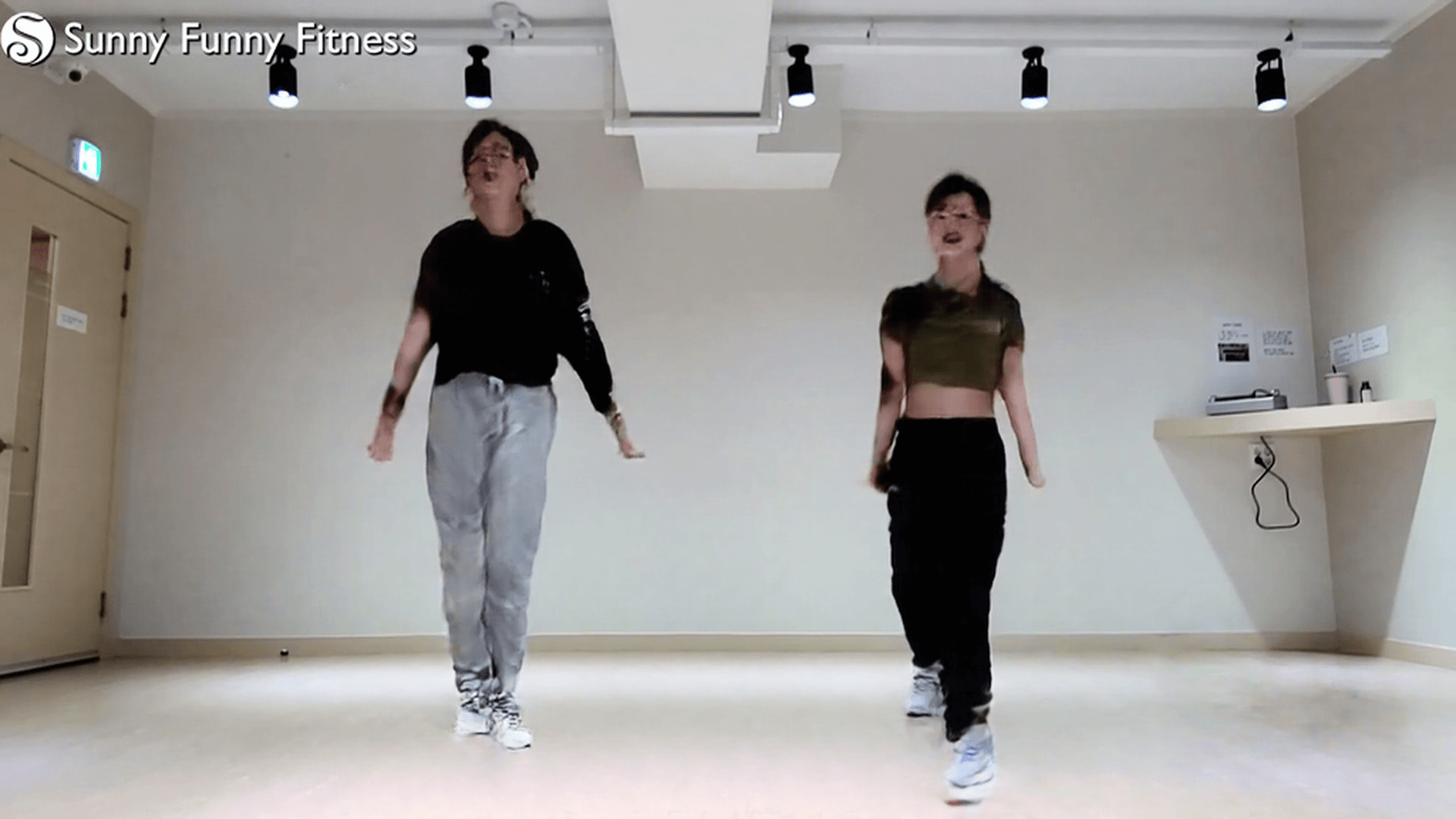}
	\includegraphics[width=0.32\textwidth]{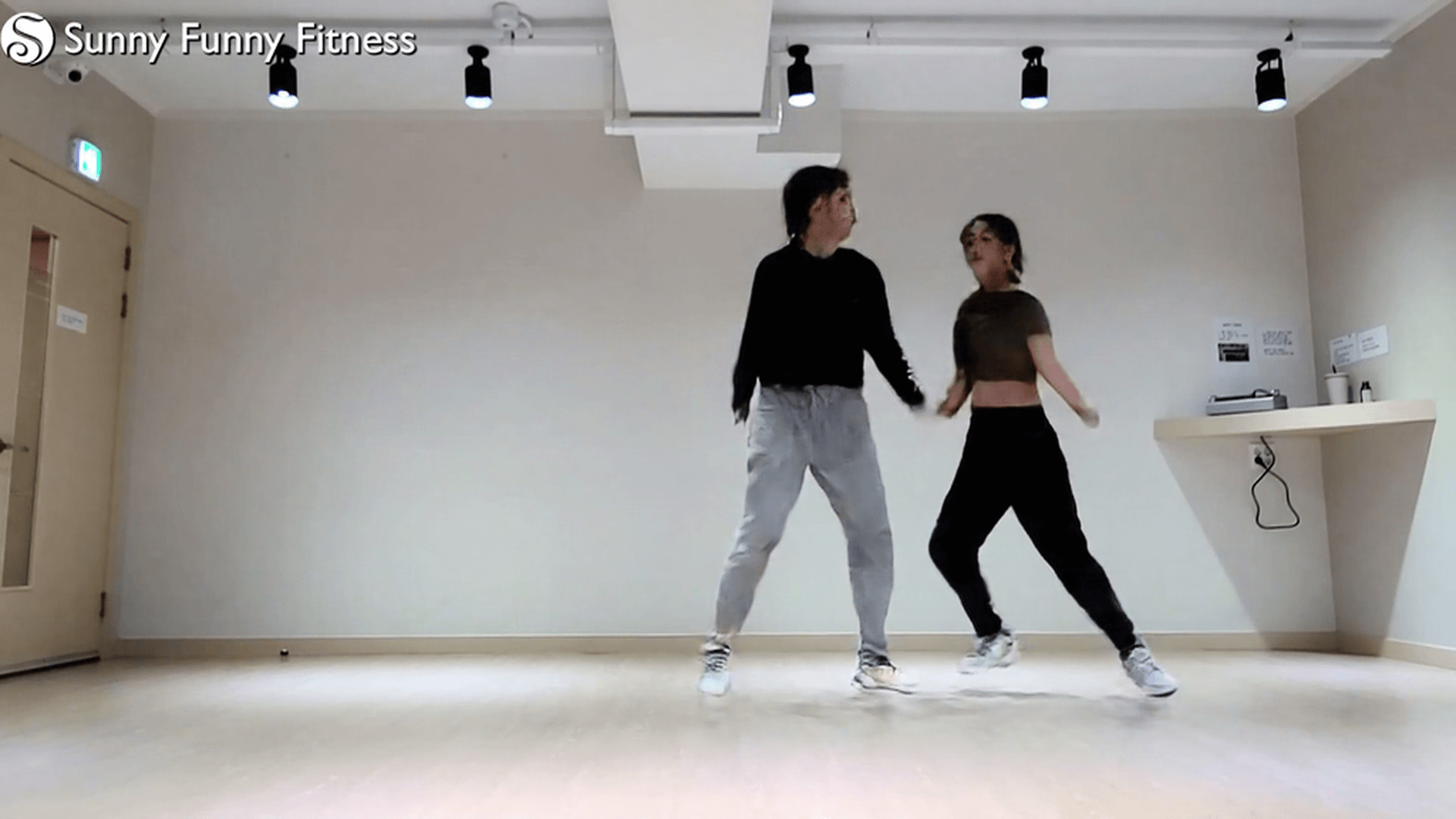}
	\includegraphics[width=0.32\textwidth]{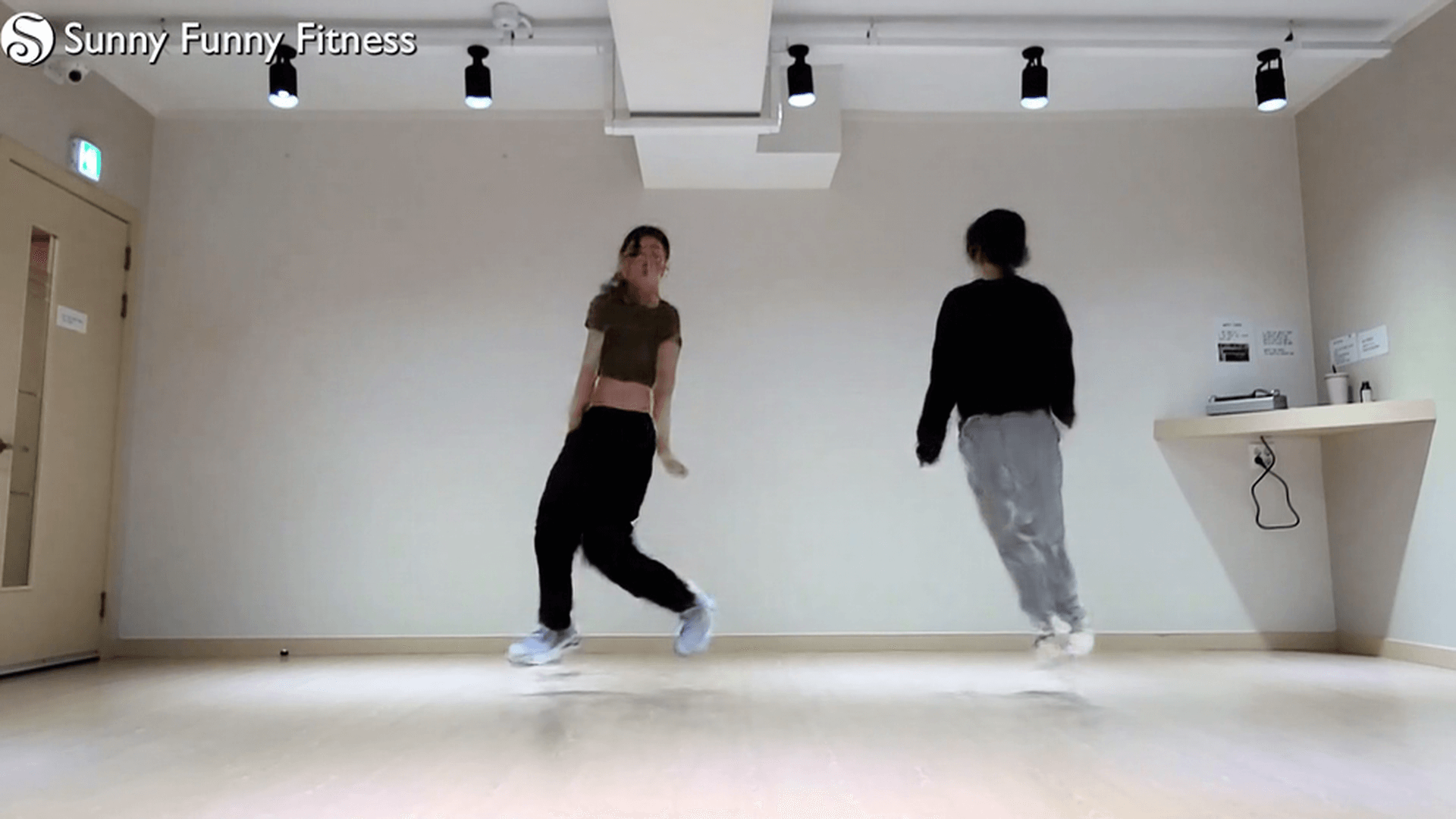}

}
\caption{Transfer results on~\cite{2person2} (bottom) with input targets from~\cite{2person1} (top) switching places without identity switch.}
\label{fig:results:no-switch}
\end{figure*}

\begin{figure*}
    \centering
\subfloat{
    \includegraphics[width=0.48\textwidth]{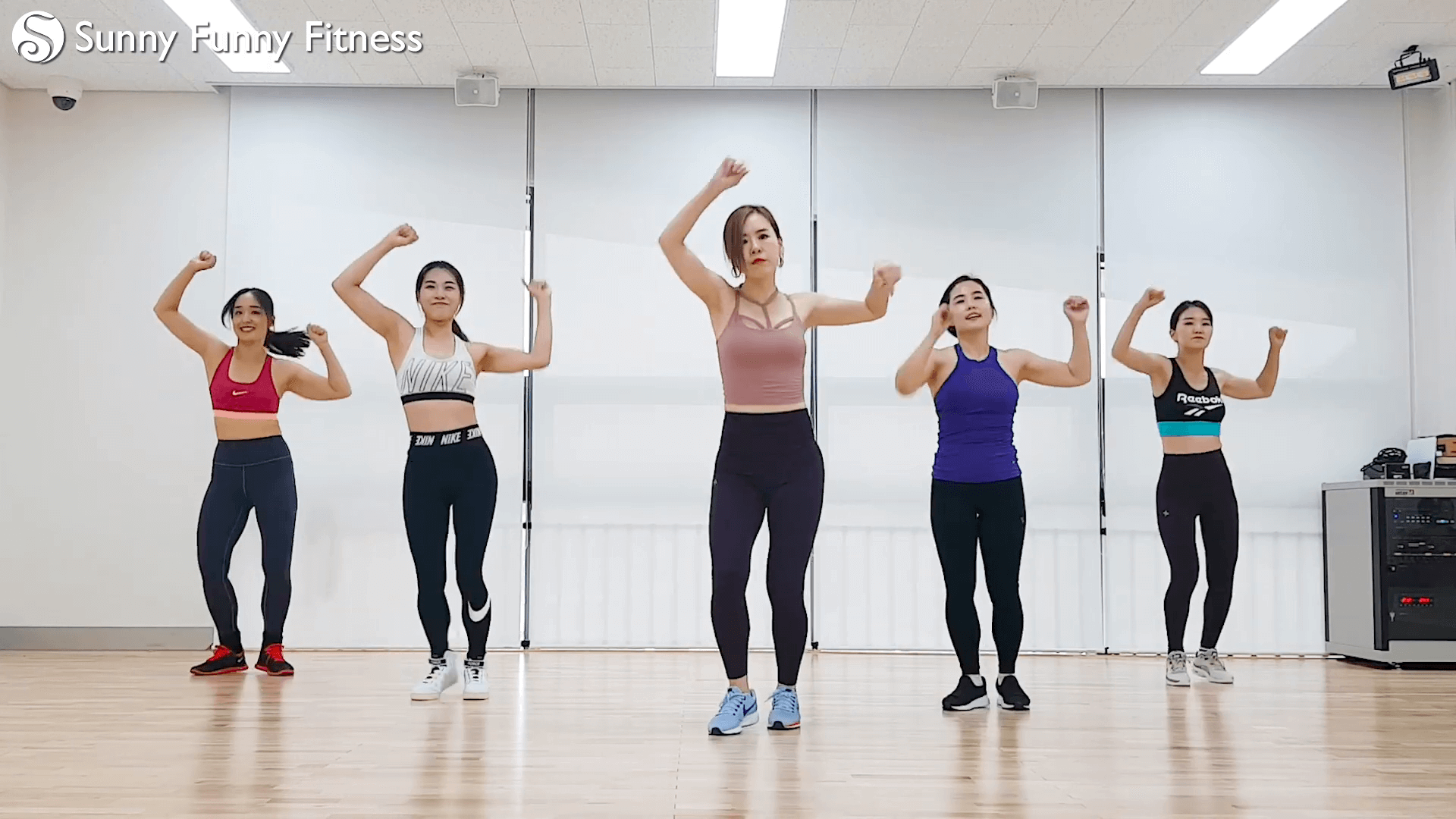}
    \includegraphics[width=0.48\textwidth]{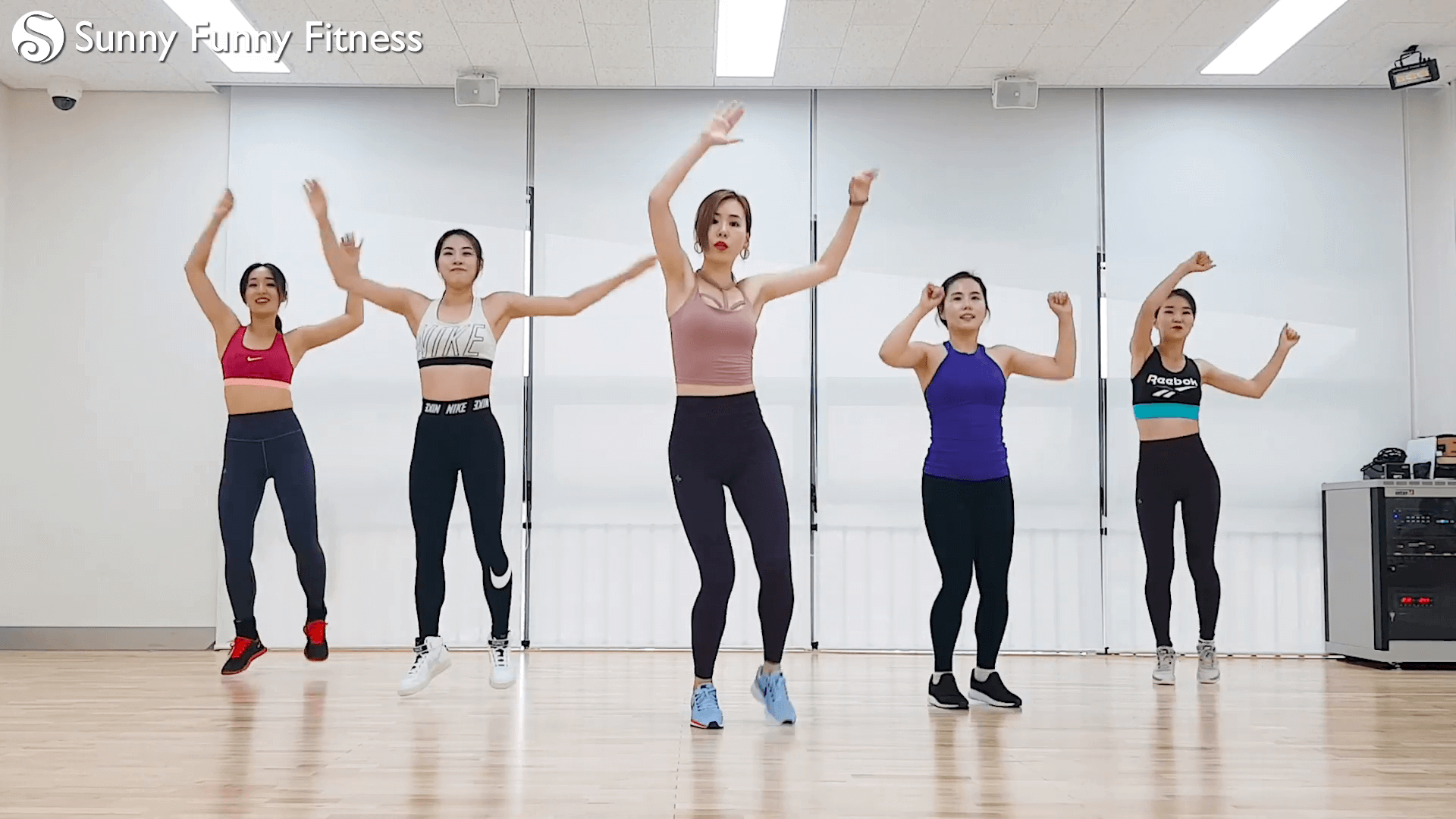}
    }\qquad
\subfloat{
    \includegraphics[width=0.48\textwidth]{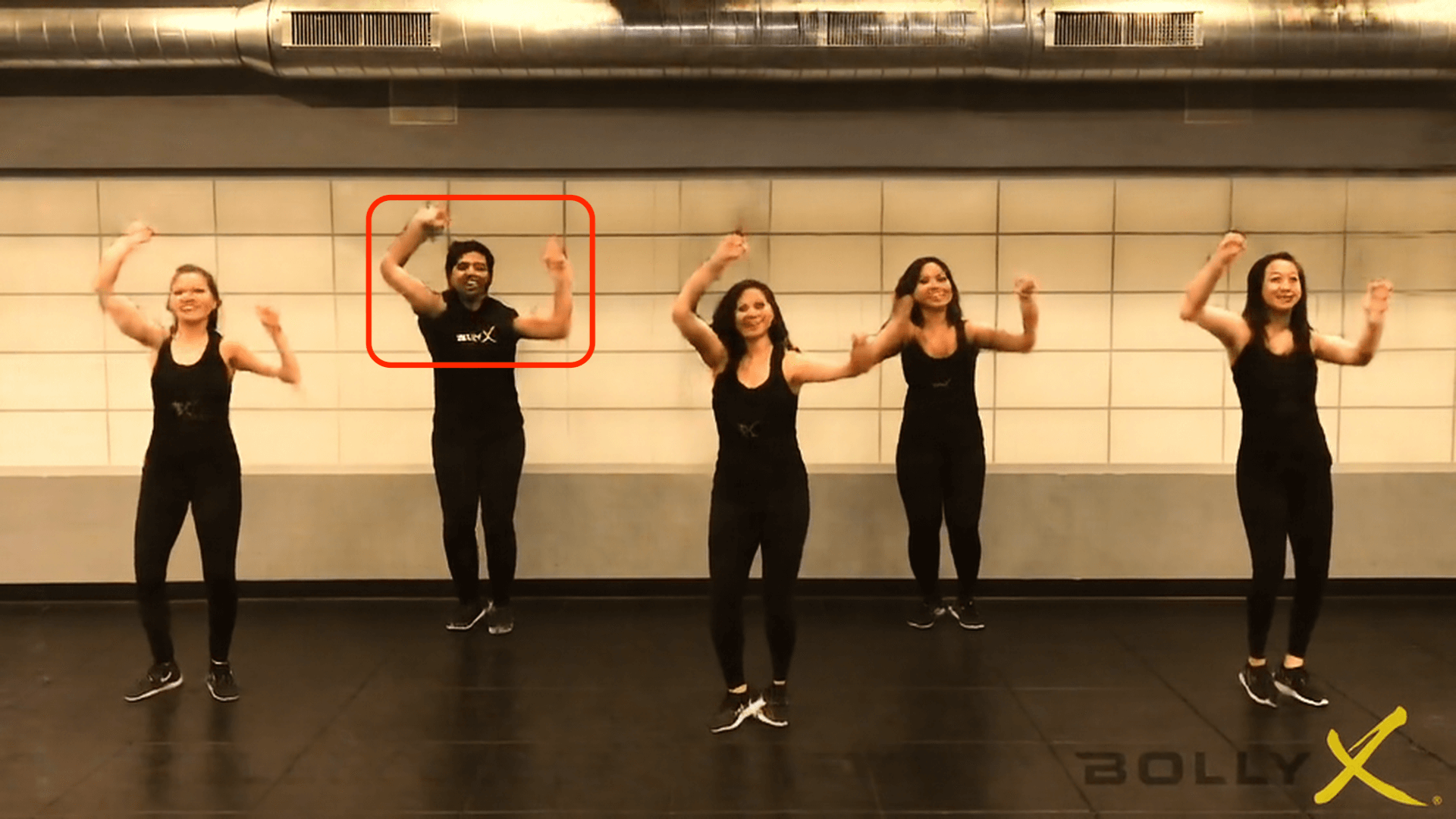}
        \includegraphics[width=0.48\textwidth]{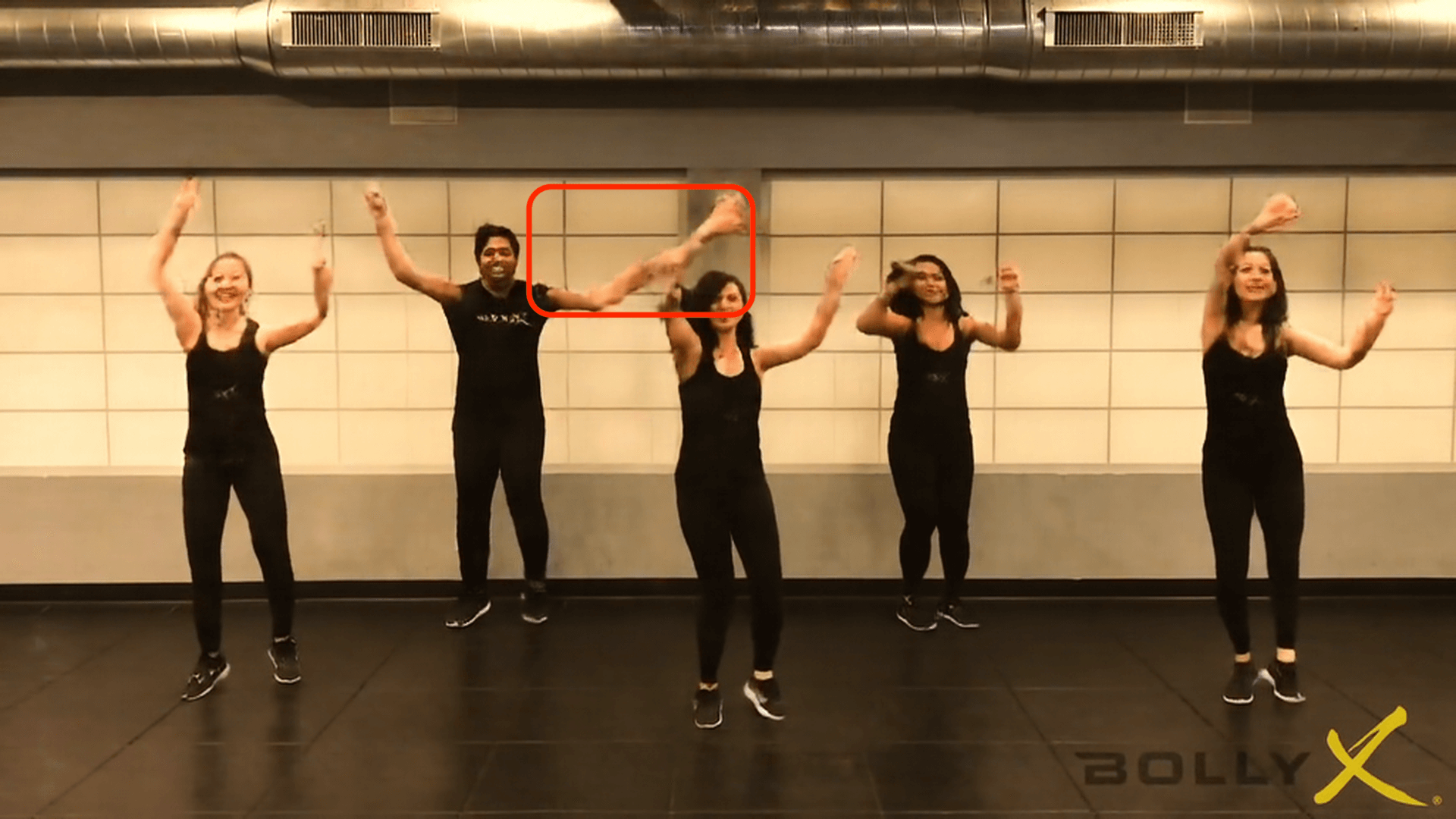}

    }
    \caption{Failure cases. Input appearance from~\cite{5person1} (top) followed by our results (bottom) on~\cite{5person2}. If our model has seen only relatively small motions for training, it can't handle extreme poses.}
    \label{fig:results:fails_extrem_poses}
\end{figure*}

Transfer results for multiple source and target subjects can be seen in Figure~\ref{fig:transfer_1} and Figure~\ref{fig:transfer_2}.
The advantage of using target normalization can be clearly seen in Figure~\ref{fig:transfer_1} where the input subjects are shifted to the left as in the learned target video. This property is important since the target scene could contain physical objects on which the person would otherwise mistakenly be projected. We show results for more difficult face poses in Figure~\ref{fig:transfer_2} with an example for which a turning face is handled properly and another for which our model struggles. In this case the turning head of the dancer in the input video has never been seen in a similar fashion during training for the target subject. Therefore, our model can't handle the projection of the back of the head and produces a strong artifact instead of generating hair.  Furthermore, failure cases for previously unseen extreme poses are illustrated in Figure~\ref{fig:results:fails_extrem_poses}. 
As shown in Table~\ref{tab:qualitative_eval_no-face}, the number of subjects for transfer grows, the performance of the model generally declines, which could be expected while increasing the difficulty of the task without altering the parameters of the model. However, the 5-to-5 model delivers contradicting results. We show qualitative results for this model in Figure~\ref{fig:results:5p_good}. Those results and particularly the faces are convincingly smooth. We notice the similarities in clothing between the target subjects. This setup not only boosts the performance metric for this video due to the clothing, but also allows better performance for the whole scene. We argue that this is related to the limited amount of parameters available to our model: less parameters are required to learn the appearances of lower bodies, therefore more parameters are available to accurately represent faces and upper bodies. Further works could progressively increase the number of subjects and investigate a required size of the model in order to reach optimal transfer performance. 
Finally, identity switches are handled as shown in Figure~\ref{fig:results:no-switch}. However, such tracking is highly dependent on the quality of the pose estimator. Therefore, for input scenes in which a subject disappears for a long time behind another subject, the track may be lost requiring a new mapping from source to target.
While our results suggest the clear feasibility to convincingly transfer multiple persons at the same time, we find that the quality of the synthesized face still needs improvement. Furthermore, extreme arm or face poses remain challenging.

\section{Conclusion}

We extended and generalized the concept of video human motion transfer to multiple person using a relatively simple but yet efficient model. We address the pose normalization of multiple subjects and potential identity switches when different actors change places. Our method, while using only a few thousand frames, delivers high-quality videos of a target group of persons following the visual instructions of another group, even generating convincing shadows. However, our results are highly limited by the available data for the target group, which is difficult to collect. Furthermore, input and target videos are required to take on a similar perspective. Future work could focus on the training data and on extracting even more information such as semantic masks, dense poses or clothing information. A potential application is to create photo-realistic avatars from synthesized poses in order to efficiently render individuals anonymous and therefore facilitate a generation of new realistic data in the target domain.

{\small
\bibliographystyle{ieee_fullname}
\bibliography{refs}
}
\vspace{12pt}
\color{red}

\end{document}